\documentclass[5p,twocolumn,authoryear,preprint]{elsarticle}
\usepackage{stfloats}
\usepackage{graphicx}
\usepackage[dvipsnames,table]{xcolor}
\usepackage{longtable}
\usepackage{flafter}
\usepackage{amsmath}
\usepackage{amssymb}
\usepackage{tabularx}
\usepackage{pdfpages}
\usepackage{subcaption}
\usepackage{stmaryrd}
\usepackage{float}
\usepackage{array}
\usepackage[nottoc,notlof,notlot]{tocbibind}
\usepackage{xurl}
\usepackage{booktabs}
\usepackage{tikz}
\usepackage{tikz-3dplot}
\usepackage{pgfplots}
\usepackage{pgfplotstable}
\usepackage{filecontents}
\usepackage{mathrsfs}
\usepackage{cuted}
\usepackage[ruled,linesnumbered]{algorithm2e}
\usepackage{tabularx}
\usepackage{subcaption}

\newcommand{\RN}[1]{\uppercase\expandafter{\romannumeral#1}}
\usepackage{setspace}
\usepackage[bottom]{footmisc} 
\usepackage{hyperref}
\usepackage{babel}
\usepackage[capitalise]{cleveref}
\usepackage{tikz}
\usepackage{ifthen}
\usetikzlibrary{calc}
\usepackage{intcalc}
\usepackage{bigints} 
\usepackage{neuralnetwork} 
\usepackage{amsmath,mathtools}
\usetikzlibrary{positioning}
\usetikzlibrary{arrows}
\usepackage{blkarray, bigstrut} 

\usepackage[parfill]{parskip}

\usepackage{pgfplots}
  \pgfplotsset{compat=newest}
  \usetikzlibrary{plotmarks}
  \usetikzlibrary{arrows.meta}
  \usepgfplotslibrary{patchplots}
  \usepackage{grffile}
  \usepackage{amsmath}

  \pgfplotsset{plot coordinates/math parser=false}
  \newlength\figureheight
  \newlength\figurewidth

\usetikzlibrary{decorations.markings}
\usetikzlibrary{decorations.pathmorphing}
\usetikzlibrary{arrows.meta}
\usetikzlibrary{shadows,arrows,positioning,shapes.geometric}
\pgfdeclarelayer{background}
\pgfdeclarelayer{foreground}
\pgfsetlayers{background,main,foreground}
\pgfplotsset{every axis/.append style={
                    label style={font=\scriptsize},
                    tick label style={font=\scriptsize}, 
                    legend style={font=\scriptsize}
                    }}

\graphicspath{{header/pictures/}{body/pictures/}{footer/pictures/}}

\AlgoDontDisplayBlockMarkers\AlgoDontDisplayGroupMarkers

\newlength{\picWidth}
\definecolor{myviolet}{rgb}{0.2, 0.2, 0.7} 

\newcommand{\x}[0]{\mathbf{x}}

\newcommand{\R}{\mathbb{R}}

\usepackage{xcolor}

\begin{document}

\begin{frontmatter}
\title{Intrinsic Dimension Estimating Autoencoder (IDEA) Using CancelOut Layer and a Projected Loss\\
        \small Application to Vertically Resolved Shallow Flow Simulations}

\author[1,2]{Antoine Oriou \corref{cor1}}
\ead{antoine.oriou@student-cs.fr}
\author[3]{Philipp Krah}
\ead{philipp.krah@cea.fr}
\author[1,4]{Julian Koellermeier}
\ead{julian.koellermeier@ugent.be}

\cortext[cor1]{Corresponding author}
\address[1]{Department of Mathematics, Computer Science and Statistics, University of Ghent, Belgium}
\address[2]{CentraleSupélec, University Paris-Saclay, France}
\address[3]{IRFM, CEA Cadarache, France}
\address[4]{Bernoulli Institute, University of Groningen, Netherlands}

\begin{abstract}
This paper introduces the Intrinsic Dimension Estimating Autoencoder (IDEA), which identifies the underlying intrinsic dimension of a wide range of datasets whose samples lie on either linear or nonlinear manifolds. 
Beyond estimating the intrinsic dimension, IDEA is also able to reconstruct the original dataset after projecting it onto the corresponding latent space, which is structured using re-weighted double CancelOut layers. Our key contribution is the introduction of the \textit{projected reconstruction loss} term, guiding the training of the model by continuously assessing the reconstruction quality under the removal of an additional latent dimension. \\
We first assess the performance of IDEA on a series of theoretical benchmarks to validate its robustness. 
These experiments allow us to test its reconstruction ability and compare its performance with state-of-the-art intrinsic dimension estimators. The benchmarks show good accuracy and high versatility of our approach.\\
Subsequently, we apply our model to data generated from the numerical solution of a vertically resolved one-dimensional free-surface flow, following a pointwise discretization of the vertical velocity profile in the horizontal direction, vertical direction, and time. IDEA succeeds in estimating the dataset’s intrinsic dimension and then reconstructs the original solution by working directly within the projection space identified by the network.
\end{abstract}
\end{frontmatter}


\section{Introduction}

The intrinsic dimension (ID) $d$ of a dataset is a fundamental quantity that refers to the minimum number of parameters required to represent the data with satisfactory accuracy (\cite{Bac_2021}). 
We consider datasets represented as matrices \(M \in \mathcal{M}_{n,p}(\R)\) where, following standard machine learning notation, \(n\) denotes the number of samples and \(p\) the number of features.
Such a dataset corresponds to a collection of points in \(\R^p\), where \(p\) is the embedded dimension. 

The objective is to determine the intrinsic dimension \( d < p \) of such datasets and to accurately reconstruct the data using only $d$ parameters.

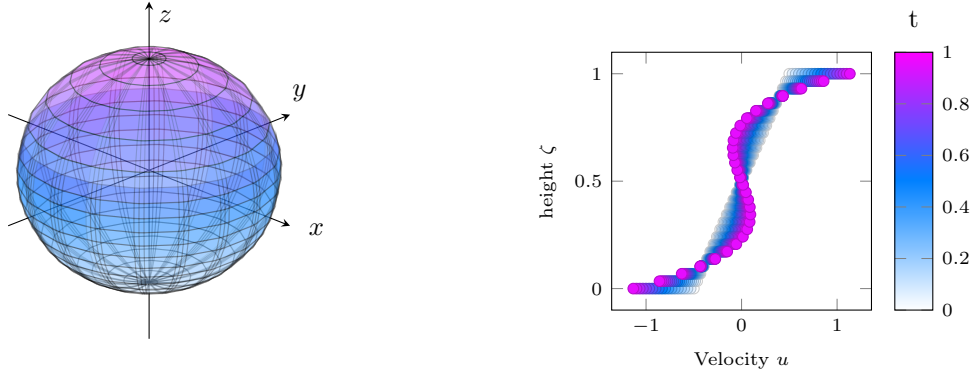
\begin{figure*}[htbp!]
    \centering
    \begin{subfigure}[t]{\picWidth}
        \centering
        \raisebox{5mm}{\begin{tikzpicture}
  \begin{axis}[
    view={45}{25},
    axis lines=center,
    ticks=none,
    xlabel={$x$}, xlabel style={at={(axis cs:1.6,0,0)},anchor=west},
    ylabel={$y$}, ylabel style={at={(axis cs:0,1.6,0)},anchor=south},
    zlabel={$z$},
    enlargelimits=false,
    xmin=-1.5, xmax=1.5,
    ymin=-1.5, ymax=1.5,
    zmin=-1.5, zmax=1.5,
    xtick=\empty, ytick=\empty, ztick=\empty,
    width=9cm, height=9cm,
    colormap/cool,
  ]

    \addplot3[
      surf,
      opacity=0.1,
      shader=flat,
      samples=40,
      samples y=30,
      z buffer=sort,
      draw=black,
      mesh/interior colormap name=cool,
    ]
    (
      {sin(deg(x))*cos(deg(y))},
      {sin(deg(x))*sin(deg(y))},
      {cos(deg(x))}
    );

  \end{axis}
\end{tikzpicture}}  
        \caption{2-dimensional sphere embedded in 3-dimensional space ($p = 3$ and $d = 2$).}
        \label{fig:3dSphere}
    \end{subfigure}
    ~
    \begin{subfigure}[t]{\picWidth}
        \centering
        \pgfplotsset{compat=1.18}

\begin{tikzpicture}
  \begin{axis}[
    width=5cm,                     
    height=5cm,                      
    xlabel={Velocity $u$},              
    ylabel={height $\zeta$},             
    colorbar,                       
    colorbar style={
      title={t},               
    },
    point meta min=0,               
    point meta max=1,               
    colormap/cool,               
  ]

    \addplot[
      scatter,
      only marks,
      mark=*,
      scatter src=explicit,         
    ]
    table[
      col sep=comma,
      x=y,
      y=x,
      meta=variable           
    ] {body/figures/leg1.csv};

  \end{axis}
\end{tikzpicture}
        \caption{20 samples taken from a 1-intrinsic-dimensional dataset embedded in a 30 dimensional space ($p = 30$ and $d = 1$).}
        \label{fig:exemple_samples}
    \end{subfigure}
    \caption{Example datasets where the intrinsic dimension of the sampling space differs from its embedded dimension.}
    \label{fig:examples_datasets}
\end{figure*}

We illustrate this distinction with two contrasting examples where the intrinsic and embedded dimensions differ, shown in Fig. \ref{fig:examples_datasets}.\\

In Fig. \ref{fig:3dSphere}, the dataset consists of $n$ sample points $(x_i, y_i, z_i)$ such that $x_i^2 + y_i^2 + z_i^2 = 1$ $\forall i \in \{1, \ldots, n\}$, describing points on a 2D sphere embedded in 3D space. 
Despite being represented in three dimensions, the data lies on a 2D surface. A well-designed model should infer that two parameters, such as azimuthal and polar angles, are sufficient to reconstruct the data without loss.

Fig. \ref{fig:exemple_samples} displays 20 samples from a dataset of $n$ sampled velocity profiles $u_{t_i}(\zeta)$ evaluated over a 30-point grid $\zeta_j\in \left[ 0, 1 \right]$ on vertical axis $\forall i \in \{1, \ldots, n\}$. Although each sample appears as a point in 30D space (one feature per grid evaluation), the underlying generative process is governed by a low-dimensional parameterization:
\begin{equation}
    t \mapsto u(\zeta)_t = - \frac{1}{2} \tilde{P}_1(\zeta) - \frac{t}{30} \tilde{P}_3(\zeta), \quad \forall t \in \left[ 0, 1 \right], 
\end{equation}
where $\tilde{P}_n(\zeta)$ denotes the scaled Legendre polynomial of degree $n$ as defined in \cite{TorrilhonKowalski}. If the model can learn the single generative parameter $t$ for each profile, it should, in principle, reconstruct the full 30-dimensional signal with negligible loss, revealing the dataset’s intrinsic dimension to be $d=1$.
This second example is of particular interest, as it closely resembles our target application: the analysis of velocity profiles of the numerical solution of vertically resolved one-dimensional free-surface flow, see \cite{TorrilhonKowalski}. 

Intrinsic dimension estimation methods have been widely used in different fields such as biology (\cite{bio}), finance (\cite{sirignano2016deeplearninglimitorder}), and physics (\cite{lee2019modelreductiondynamicalsystems}). Our application addresses the efficient numerical solution of a vertically resolved one-dimensional
free-surface flow.  Similar to the work of \cite{lee2019modelreductiondynamicalsystems}, we employ deep learning methods to overcome the loss of accuracy inherent in linear model reduction techniques such as Proper Orthogonal Decomposition (POD).

Over the years, numerous methods have been proposed to estimate the intrinsic dimension of data (\cite{article},  \cite{binnie2025surveydimensionestimationmethods}), often classified by the geometric information they exploit: tangential estimators (PCA \cite{fan2010intrinsicdimensionestimationdata}) detecting local affine structure, parametric estimators based on dimension-dependent probability distributions (Correlation Integral \cite{GRASSBERGER1983189}, and topological or metric-based estimators. 
IDEA differs from these approaches by enabling (in addition to the estimation of $d$) the reconstruction of the data after the compression in a $d$-dimensional space.
It leverages a flexible implementation of controlled sparsity in a neural network double CancelOut layer, adapted from \cite{inbook}. Our main contribution lies in the loss function and training procedure, which enforce strong regularization and enable intrinsic dimension estimation through the latent space. 
Common challenges for such estimators include sensitivity to hyperparameters, large sample requirements, poor performance in high dimensions, and difficulty handling non-linear geometries (\cite{article}). Unlike classical methods, IDEA addresses many of these limitations and does not suffer from explicit weakness. It only requires a sufficiently large sample size, owing to its data-driven nature and reconstruction capabilities.

The rest of the paper is organized as follows. In \cref{sec:2}, we explain the structure of IDEA, detailing the training loop and the latent space regularization process. In \cref{sec:03}, we assess the performance of IDEA on synthetic data. In \cref{sec:4}, we compare IDEA with other estimators on benchmark datasets from \cite{article}. Finally, in \cref{sec:5}, we apply IDEA to numerical data stemming from one-dimensional free-surface flow simulations from \cite{TorrilhonKowalski} and showcase the efficiency of our approach to reduce high-dimensional data from numerical solutions to their intrinsic dimension. The paper ends with concluding remarks.

\section{Mathematical background}
\label{sec:2}
\subsection{Autoencoder neural network}

An autoencoder \cite{rumelhart:errorpropnonote} is a function that aims to compress and reconstruct an input while minimizing the reconstruction error. It has been widely used for diverse machine learning tasks such as image segmentation, classification, and time series predictions, among others \cite{michelucci2022introductionautoencoders}. 
An autoencoder is typically divided into three parts: \textit{encoder}, \textit{bottleneck}, and \textit{decoder}. The \textit{encoder} is a function $f_\theta : \mathbb{R}^p \rightarrow \mathbb{R}^l$, with $l \ll p$, which maps the input data to a lower-dimensional latent space, the \textit{bottleneck}.
The \textit{decoder} is a function $g_\phi : \mathbb{R}^l \rightarrow \mathbb{R}^p$, which reconstructs the input from the latent representation.

In the context of intrinsic dimension estimation, the structure of the latent space $\mathbb{R}^l$ plays a central role, as it provides an upper bound of the ID under sufficient reconstruction precision.

\subsection{Structure of the model}

We introduce the Intrinsic Dimension Estimating Autoencoder (IDEA), a neural network autoencoder specialized for intrinsic dimension estimation (see \cite{Goodfellow-et-al-2016} for details about neural network implementations).

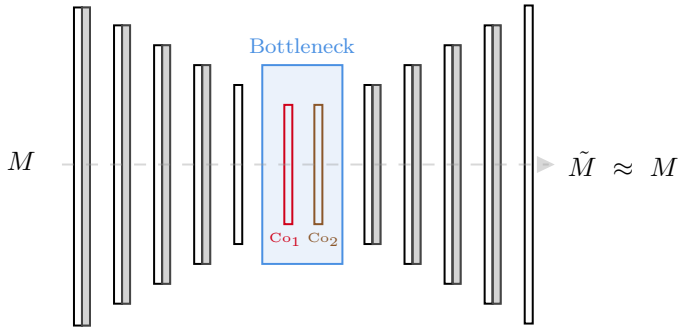
\begin{figure}[htp!]
	\setlength\figureheight{0.2\textwidth}
	\setlength\figurewidth{0.2\textwidth}
	\centering
    \tikzset{every picture/.style={line width=0.75pt}} 

\begin{tikzpicture}[x=0.75pt,y=0.75pt,yscale=-1,xscale=1]

\draw   (166,60) -- (170,60) -- (170,200) -- (166,200) -- cycle ;
\draw   (186,70) -- (190,70) -- (190,190) -- (186,190) -- cycle ;
\draw   (206,80) -- (210,80) -- (210,180) -- (206,180) -- cycle ;
\draw   (226,90) -- (230,90) -- (230,170) -- (226,170) -- cycle ;
\draw   (170,60) -- (174,60) -- (174,200) -- (170,200) -- cycle ;
\draw   (190,70) -- (194,70) -- (194,190) -- (190,190) -- cycle ;
\draw   (210,80) -- (214,80) -- (214,180) -- (210,180) -- cycle ;
\draw   (371,50) -- (375,50) -- (375,210) -- (371,210) -- cycle ;
\draw   (351,60) -- (355,60) -- (355,200) -- (351,200) -- cycle ;
\draw   (331,70) -- (335,70) -- (335,190) -- (331,190) -- cycle ;
\draw   (311,80) -- (315,80) -- (315,180) -- (311,180) -- cycle ;
\draw   (291,90) -- (295,90) -- (295,170) -- (291,170) -- cycle ;
\draw   (355,60) -- (359,60) -- (359,200) -- (355,200) -- cycle ;
\draw   (335,70) -- (339,70) -- (339,190) -- (335,190) -- cycle ;
\draw   (315,80) -- (319,80) -- (319,180) -- (315,180) -- cycle ;
\draw   (146,51) -- (150,51) -- (150,211) -- (146,211) -- cycle ;
\draw  [color={rgb, 255:red, 155; green, 155; blue, 155 }  ,draw opacity=0.45 ][fill={rgb, 255:red, 155; green, 155; blue, 155 }  ,fill opacity=0.43 ] (170,60) -- (174,60) -- (174,200) -- (170,200) -- cycle ;
\draw  [color={rgb, 255:red, 155; green, 155; blue, 155 }  ,draw opacity=0.45 ][fill={rgb, 255:red, 155; green, 155; blue, 155 }  ,fill opacity=0.43 ] (190,70) -- (194,70) -- (194,190) -- (190,190) -- cycle ;
\draw  [color={rgb, 255:red, 155; green, 155; blue, 155 }  ,draw opacity=0.45 ][fill={rgb, 255:red, 155; green, 155; blue, 155 }  ,fill opacity=0.43 ] (210,80) -- (214,80) -- (214,180) -- (210,180) -- cycle ;
\draw  [color={rgb, 255:red, 155; green, 155; blue, 155 }  ,draw opacity=0.45 ][fill={rgb, 255:red, 155; green, 155; blue, 155 }  ,fill opacity=0.43 ] (315,80) -- (319,80) -- (319,180) -- (315,180) -- cycle ;
\draw  [color={rgb, 255:red, 155; green, 155; blue, 155 }  ,draw opacity=0.45 ][fill={rgb, 255:red, 155; green, 155; blue, 155 }  ,fill opacity=0.43 ] (335,70) -- (339,70) -- (339,190) -- (335,190) -- cycle ;
\draw  [color={rgb, 255:red, 155; green, 155; blue, 155 }  ,draw opacity=0.45 ][fill={rgb, 255:red, 155; green, 155; blue, 155 }  ,fill opacity=0.43 ] (355,60) -- (359,60) -- (359,200) -- (355,200) -- cycle ;
\draw   (150,51) -- (154,51) -- (154,211) -- (150,211) -- cycle ;
\draw  [color={rgb, 255:red, 155; green, 155; blue, 155 }  ,draw opacity=0.45 ][fill={rgb, 255:red, 155; green, 155; blue, 155 }  ,fill opacity=0.43 ] (150,51) -- (154,51) -- (154,211) -- (150,211) -- cycle ;
\draw  [color={rgb, 255:red, 74; green, 144; blue, 226 }  ,draw opacity=1 ][fill={rgb, 255:red, 74; green, 144; blue, 226 }  ,fill opacity=0.11 ] (240,80) -- (280,80) -- (280,180) -- (240,180) -- cycle ;
\draw  [color={rgb, 255:red, 208; green, 2; blue, 27 }  ,draw opacity=1 ] (251,100) -- (255,100) -- (255,160) -- (251,160) -- cycle ;
\draw  [color={rgb, 255:red, 139; green, 87; blue, 42 }  ,draw opacity=1 ] (266,100) -- (270,100) -- (270,160) -- (266,160) -- cycle ;
\draw   (295,90) -- (299,90) -- (299,170) -- (295,170) -- cycle ;
\draw  [color={rgb, 255:red, 155; green, 155; blue, 155 }  ,draw opacity=0.45 ][fill={rgb, 255:red, 155; green, 155; blue, 155 }  ,fill opacity=0.43 ] (295,90) -- (299,90) -- (299,170) -- (295,170) -- cycle ;
\draw [color={rgb, 255:red, 155; green, 155; blue, 155 }  ,draw opacity=0.39 ] [dash pattern={on 4.5pt off 4.5pt}]  (140,130) -- (383,130) ;
\draw [shift={(386,130)}, rotate = 180] [fill={rgb, 255:red, 155; green, 155; blue, 155 }  ,fill opacity=0.39 ][line width=0.08]  [draw opacity=0] (8.93,-4.29) -- (0,0) -- (8.93,4.29) -- cycle    ;

\draw (232,65) node [anchor=north west][inner sep=0.75pt]  [color={rgb, 255:red, 74; green, 144; blue, 226 }  ,opacity=1 ] [align=left] {\footnotesize Bottleneck};
\draw (242,163) node [anchor=north west][inner sep=0.75pt]  [color={rgb, 255:red, 208; green, 2; blue, 27 }  ,opacity=1 ] [align=left] {{\tiny Co$\displaystyle _{1}$}};
\draw (261,163) node [anchor=north west][inner sep=0.75pt]  [color={rgb, 255:red, 139; green, 87; blue, 42 }  ,opacity=1 ] [align=left] {{\tiny Co$\displaystyle _{2}$}};
\draw (111,122) node [anchor=north west][inner sep=0.75pt]   [align=left] {$\displaystyle M$};
\draw (391,122) node [anchor=north west][inner sep=0.75pt]   [align=left] {$\displaystyle \tilde{M} \ \approx \ M$};

\end{tikzpicture}
    \caption{IDEA's architecture consisting of an encoder, a re-weighted double CancelOut (Co) bottleneck, and a decoder. Grey rectangles represent a block composed of a normalization layer followed by a SiLU activation. The layer sizes are set as follows: $128 \rightarrow 64 \rightarrow 32 \rightarrow 16 \rightarrow {\color[HTML]{5395e4}l \rightarrow l} \color{black} \rightarrow 16 \rightarrow 32 \rightarrow 64 \rightarrow 128$.}
    \label{fig:IDEA}
\end{figure}

IDEA's encoder and decoder are each composed of five fully connected linear layers that map the input to the latent space and vice versa (see Fig. \ref{fig:IDEA}). Each layer (except the last) is followed by a normalization layer and a SiLU activation function (respectively introduced in \cite{ba2016layernormalization} and \cite{elfwing2017sigmoidweightedlinearunitsneural}).

To promote dimensionality reduction, the latent space is structured to encourage the elimination of redundant variables. A key component enabling this is the CancelOut layer, here adapted from \cite{inbook}. This specialized neural network layer selects a subset of informative latent variables by suppressing those that do not contribute meaningfully to the reconstruction (see Fig. ~\ref{fig:CO}). Given the central role of the CancelOut layer in our architecture, we provide detailed implementation insights in the following Section~\ref{subsec:cancelout}. 

\subsection{Re-weighted double CancelOut layers} 
\label{subsec:cancelout}

Formally, we refer to the bottleneck component as the \textit{Re-weighted Double CancelOut Layer} as we constructed the bottleneck as a superposition of two CancelOut layers with distinct objectives to facilitate the regularization process. We begin by briefly recalling the functioning of the CancelOut layer. \\
The CancelOut operation is defined as:
\begin{equation}
\text{CancelOut}(X) = X \odot g(W_{\mathrm{CO}}),
\end{equation}
where $\odot$ denotes element-wise multiplication, $X \in \mathbb{R}^{l}$ is the input vector, $W_{\mathrm{CO}} \in \mathbb{R}^{l}$ is a learned weight vector, and $g$ is an activation function (ReLU) applied element-wise.

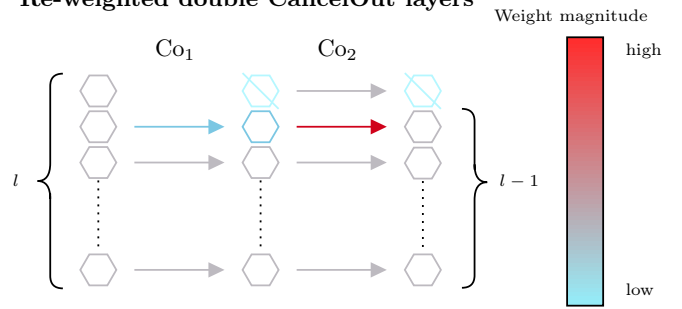
\begin{figure}[h!]
	\setlength\figureheight{0.20\textwidth}
	\setlength\figurewidth{0.20\textwidth}
	\centering
    \scalebox{0.9}{
  
\tikzset {_ciiu4xm0c/.code = {\pgfsetadditionalshadetransform{ \pgftransformshift{\pgfpoint{0 bp } { 0 bp }  }  \pgftransformrotate{-90 }  \pgftransformscale{2 }  }}}
\pgfdeclarehorizontalshading{_tvqh1ubnr}{150bp}{rgb(0bp)=(1,0.18,0.18);
rgb(37.5bp)=(1,0.18,0.18);
rgb(62.5bp)=(0.57,0.95,1);
rgb(100bp)=(0.57,0.95,1)}
\tikzset{every picture/.style={line width=0.75pt}} 

\begin{tikzpicture}[x=0.75pt,y=0.75pt,yscale=-1,xscale=1]

\draw [color={rgb, 255:red, 123; green, 199; blue, 227 }  ,draw opacity=1 ]   (180,90) -- (227,90) ;
\draw [shift={(230,90)}, rotate = 180] [fill={rgb, 255:red, 123; green, 199; blue, 227 }  ,fill opacity=1 ][line width=0.08]  [draw opacity=0] (8.93,-4.29) -- (0,0) -- (8.93,4.29) -- cycle    ;
\draw [color={rgb, 255:red, 189; green, 186; blue, 192 }  ,draw opacity=1 ]   (180,110) -- (227,110) ;
\draw [shift={(230,110)}, rotate = 180] [fill={rgb, 255:red, 189; green, 186; blue, 192 }  ,fill opacity=1 ][line width=0.08]  [draw opacity=0] (8.93,-4.29) -- (0,0) -- (8.93,4.29) -- cycle    ;
\draw [color={rgb, 255:red, 208; green, 2; blue, 27 }  ,draw opacity=1 ]   (270,90) -- (317,90) ;
\draw [shift={(320,90)}, rotate = 180] [fill={rgb, 255:red, 208; green, 2; blue, 27 }  ,fill opacity=1 ][line width=0.08]  [draw opacity=0] (8.93,-4.29) -- (0,0) -- (8.93,4.29) -- cycle    ;
\path  [shading=_tvqh1ubnr,_ciiu4xm0c] (420,40) -- (440,40) -- (440,190) -- (420,190) -- cycle ; 
 \draw   (420,40) -- (440,40) -- (440,190) -- (420,190) -- cycle ; 

\draw  [color={rgb, 255:red, 181; green, 247; blue, 255 }  ,draw opacity=1 ] (260,70) -- (255,78.66) -- (245,78.66) -- (240,70) -- (245,61.34) -- (255,61.34) -- cycle ;
\draw  [color={rgb, 255:red, 123; green, 199; blue, 227 }  ,draw opacity=1 ] (260,90) -- (255,98.66) -- (245,98.66) -- (240,90) -- (245,81.34) -- (255,81.34) -- cycle ;
\draw  [color={rgb, 255:red, 189; green, 186; blue, 192 }  ,draw opacity=1 ] (260,170) -- (255,178.66) -- (245,178.66) -- (240,170) -- (245,161.34) -- (255,161.34) -- cycle ;
\draw  [dash pattern={on 0.84pt off 2.51pt}]  (250,120) -- (250,160) ;
\draw [color={rgb, 255:red, 181; green, 247; blue, 255 }  ,draw opacity=1 ]   (240,60) -- (260,80) ;
\draw  [dash pattern={on 0.84pt off 2.51pt}]  (340,120.34) -- (340,160.34) ;
\draw  [color={rgb, 255:red, 181; green, 247; blue, 255 }  ,draw opacity=1 ] (350,70) -- (345,78.66) -- (335,78.66) -- (330,70) -- (335,61.34) -- (345,61.34) -- cycle ;
\draw [color={rgb, 255:red, 181; green, 247; blue, 255 }  ,draw opacity=1 ]   (330,60) -- (350,80) ;
\draw  [dash pattern={on 0.84pt off 2.51pt}]  (160,120) -- (160,160) ;
\draw   (140.33,60.33) .. controls (135.66,60.33) and (133.33,62.66) .. (133.33,67.33) -- (133.33,110.33) .. controls (133.33,117) and (131,120.33) .. (126.33,120.33) .. controls (131,120.33) and (133.33,123.66) .. (133.33,130.33)(133.33,127.33) -- (133.33,173.33) .. controls (133.33,178) and (135.66,180.33) .. (140.33,180.33) ;
\draw   (361.67,180) .. controls (366.34,180) and (368.67,177.67) .. (368.67,173) -- (368.67,131) .. controls (368.67,124.33) and (371,121) .. (375.67,121) .. controls (371,121) and (368.67,117.67) .. (368.67,111) (368.67,114) -- (368.67,87) .. controls (368.67,82.33) and (366.34,80) .. (361.67,80) ;
\draw [color={rgb, 255:red, 189; green, 186; blue, 192 }  ,draw opacity=1 ]   (180,170) -- (227,170) ;
\draw [shift={(230,170)}, rotate = 180] [fill={rgb, 255:red, 189; green, 186; blue, 192 }  ,fill opacity=1 ][line width=0.08]  [draw opacity=0] (8.93,-4.29) -- (0,0) -- (8.93,4.29) -- cycle    ;
\draw [color={rgb, 255:red, 189; green, 186; blue, 192 }  ,draw opacity=1 ]   (270,110) -- (317,110) ;
\draw [shift={(320,110)}, rotate = 180] [fill={rgb, 255:red, 189; green, 186; blue, 192 }  ,fill opacity=1 ][line width=0.08]  [draw opacity=0] (8.93,-4.29) -- (0,0) -- (8.93,4.29) -- cycle    ;
\draw [color={rgb, 255:red, 189; green, 186; blue, 192 }  ,draw opacity=1 ]   (270,170) -- (317,170) ;
\draw [shift={(320,170)}, rotate = 180] [fill={rgb, 255:red, 189; green, 186; blue, 192 }  ,fill opacity=1 ][line width=0.08]  [draw opacity=0] (8.93,-4.29) -- (0,0) -- (8.93,4.29) -- cycle    ;
\draw  [color={rgb, 255:red, 189; green, 186; blue, 192 }  ,draw opacity=1 ] (170,170) -- (165,178.66) -- (155,178.66) -- (150,170) -- (155,161.34) -- (165,161.34) -- cycle ;
\draw  [color={rgb, 255:red, 189; green, 186; blue, 192 }  ,draw opacity=1 ] (170,110) -- (165,118.66) -- (155,118.66) -- (150,110) -- (155,101.34) -- (165,101.34) -- cycle ;
\draw  [color={rgb, 255:red, 189; green, 186; blue, 192 }  ,draw opacity=1 ] (170,90) -- (165,98.66) -- (155,98.66) -- (150,90) -- (155,81.34) -- (165,81.34) -- cycle ;
\draw  [color={rgb, 255:red, 189; green, 186; blue, 192 }  ,draw opacity=1 ] (170,70) -- (165,78.66) -- (155,78.66) -- (150,70) -- (155,61.34) -- (165,61.34) -- cycle ;
\draw  [color={rgb, 255:red, 189; green, 186; blue, 192 }  ,draw opacity=1 ] (350,90) -- (345,98.66) -- (335,98.66) -- (330,90) -- (335,81.34) -- (345,81.34) -- cycle ;
\draw  [color={rgb, 255:red, 189; green, 186; blue, 192 }  ,draw opacity=1 ] (350,110.32) -- (345,118.98) -- (335,118.98) -- (330,110.32) -- (335,101.66) -- (345,101.66) -- cycle ;
\draw  [color={rgb, 255:red, 189; green, 186; blue, 192 }  ,draw opacity=1 ] (350,170) -- (345,178.66) -- (335,178.66) -- (330,170) -- (335,161.34) -- (345,161.34) -- cycle ;
\draw  [color={rgb, 255:red, 189; green, 186; blue, 192 }  ,draw opacity=1 ] (260,110) -- (255,118.66) -- (245,118.66) -- (240,110) -- (245,101.34) -- (255,101.34) -- cycle ;
\draw [color={rgb, 255:red, 189; green, 186; blue, 192 }  ,draw opacity=1 ]   (270,70) -- (317,70) ;
\draw [shift={(320,70)}, rotate = 180] [fill={rgb, 255:red, 189; green, 186; blue, 192 }  ,fill opacity=1 ][line width=0.08]  [draw opacity=0] (8.93,-4.29) -- (0,0) -- (8.93,4.29) -- cycle    ;

\draw (111,115) node [anchor=north west][inner sep=0.75pt]  [font=\scriptsize] [align=left] {$\displaystyle l$ };
\draw (381,115) node [anchor=north west][inner sep=0.75pt]  [font=\scriptsize] [align=left] {$\displaystyle l-1$};
\draw (190,40) node [anchor=north west][inner sep=0.75pt]   [align=left] { $\displaystyle \text{Co}_{1}$};

\definecolor{myblue}{HTML}{7AC7E3} 

\draw (195,78) node[
    anchor=north west,
    inner sep=0.75pt,
    align=left,
    text=myblue,
    scale=0.6
]{};
\draw (280,40) node [anchor=north west][inner sep=0.75pt]   [align=left] {$\displaystyle \text{Co}_{2}$};
\draw (378,23) node [anchor=north west][inner sep=0.75pt]   [align=left] {\scriptsize Weight magnitude};
\draw (115,12) node [anchor=north west][inner sep=0.75pt]   [align=left] {\textbf{Re-weighted double CancelOut layers}};
\draw (451,42) node [anchor=north west][inner sep=0.75pt]   [align=left] {\scriptsize high};
\draw (451,176) node [anchor=north west][inner sep=0.75pt]   [align=left] {\scriptsize low};

\end{tikzpicture}}
    
    \caption{Re-weighted double CancelOut (Co) layers forming the bottleneck. The initial latent dimension is $l$. Weights are represented as arrows, variables as hexagons. Co$_1$ sets the last weight to zero, effectively removing one dimension from the latent space, while also beginning to regularize the current last non-zero weight (blue), diminishing the magnitude of 
    the corresponding latent variable. In response, Co$_2$ increases its corresponding weight (red) to ensure that the remaining $l-1$ variables stay on a comparable scale after the bottleneck. In the diagram, the weight regularization process unfolds from top to bottom.}
    \label{fig:CO}
\end{figure}

The CancelOut layer operation consists of a one-to-one mapping between the input latent variables and the output (See Fig. \ref{fig:CO}), where each input dimension is multiplied by a trainable scalar weight.\\
As each output variable depends only on the corresponding input variable (i.e., the one-to-one mapping ensures no mixing of information across dimensions), setting the $i^{\text{th}}$ weight to zero completely suppresses the $i^{\text{th}}$ latent variable. As a result, the corresponding dimension of the latent representation becomes inactive and no longer contributes to the reconstruction.\\
The CancelOut mechanism provides a direct and controllable path to reduce the effective dimension of the latent space: as training progresses, the model may learn to suppress uninformative or redundant latent variables. The CancelOut Layer thus acts as a sparsity-inducing gate.
Formally, the bottleneck of IDEA consists of two successive CancelOut Layers, each with 
$l$ neurons (i.e., $l$ initially non-negative weights), connected in a one-to-one fashion (see Fig. ~\ref{fig:CO}).\\
The first CancelOut layer (Co$_1$) serves as a sparsity-inducing gate: it is subject to regularization (e.g., L1), which encourages some of its weights to approach zero. This effectively removes certain latent dimensions from contributing to the output, enabling the network to learn the intrinsic dimensionality of the data. The second CancelOut Layer (Co$_2$) is also a one-to-one mapping, but it is not regularized. Its purpose is to rescale all informative features to a comparable level after the bottleneck, thereby compensating for the magnitude suppression caused by continued regularization of relevant weights.

Let $l_\mathrm{eff}$ be the number of effective variables in the latent space (number of non-zero weights of $\mathrm{Co}_1$).
If the model can successfully reconstruct the data using a bottleneck with $l_\mathrm{eff}$ effective variables, this implies that the intrinsic dimension is at most $l_\mathrm{eff}$. Our objective is therefore to encourage the model to minimize the effective dimensionality $l_\mathrm{eff}$ of the bottleneck as much as possible so that it converges towards the smallest $l_\mathrm{eff}$ preserving reconstruction quality, which is the intrinsic dimension $d$ of the dataset. This is achieved through the design of the model's loss function.

\subsection{Training loop and detailed loss function}

The training loop of our model is its main distinctive feature. We introduce a versatile loss function 
with specific terms giving the model a simple, but flexible incentive to detect the smallest $l_\mathrm{eff}$ of the dataset via regularization. This consists in adding a penalty term of the weights directly in the loss function \cite{kukačka2017regularizationdeeplearningtaxonomy}.

The regularized model is designed to reconstruct the input with the smallest $l_\mathrm{eff}$ preserving reconstruction quality, which is $d$ by definition. As $l_\mathrm{eff}$ is the number of non-zero weights of Co$_1$ (see \cref{subsec:cancelout}), we want the loss function (\cref{eq:loss_total}) to drive the Co$_1$ weights toward sparsity by promoting zero-valued entries, while keeping a good overall reconstruction loss. The loss function is 
\begin{multline}
\mathcal{L}_{\mathrm{total}} = 
\underbrace{\mathcal{L}_{\mathrm{rec}}(\hat{x}, x)}_{\mathrm{reconstruction}} + 
\underbrace{\lambda_{\mathrm{rec}} \cdot \mathcal{L}_{\mathrm{rec}}(\hat{x}_{\mathrm{adj}}, x)}_{\text{projected reconstruction}} \\+ 
\underbrace{\lambda_{\mathrm{reg}} \cdot \mathcal{L}_{\mathrm{reg}}(w_{\mathrm{co1}}^{(\text{last})}, \alpha)}_{\text{L1 regularization}} +
\underbrace{\lambda_{\mathrm{orth}} \cdot \mathcal{L}_{\mathrm{orth}}(f_{\theta}(x))}_{\text{orthogonality penalty}}\,,
\label{eq:loss_total}
\end{multline}
 where
\begin{itemize}
    \item $\hat{x} = g_{\phi}(f_{\theta}(x))$ is the main model output (full reconstruction).
    \item $\mathcal{L}_{\mathrm{rec}}$ is the MSE reconstruction loss.
    \item $w_{\mathrm{co}_1}^{(\text{last})}$ is the last non-zero weight in the Co$_1$ layer.
    \item $\lambda_{\mathrm{rec}}$, $\lambda_{\mathrm{reg}}$, $\lambda_{\mathrm{orth}}$ are weighting coefficients for projected reconstruction, regularization, and orthogonality penalty, respectively.
    \item $\hat{x}_{\mathrm{adj}}$ , $\mathcal{L}_{\mathrm{reg}}$, $\alpha$, and $\mathcal{L}_{\mathrm{orth}}$ are detailed below. 
\end{itemize}

\paragraph{Projected reconstruction: $\hat{x}_{\mathrm{adj}}$} This is the output of the model after setting the last non-zero weight of the Co$_1$ layer to zero. $\hat{x}_{\mathrm{adj}}$ is thus the reconstructed input after compression in an $l_\mathrm{eff}-1$ dimensional space.\\
While trying to reduce the reconstruction loss using $l_\mathrm{eff}$ dimensions, the model also attempts to reduce the \textit{projected reconstruction loss} $\lambda_{\mathrm{rec}} \cdot \mathcal{L}_{\mathrm{rec}}(\hat{x}_{\mathrm{adj}}, x)$, which is the loss criterion ($\mathcal{L}_{\mathrm{rec}}$)
applied to $\hat{x}_{\mathrm{adj}}$ and scaled by $\lambda_{rec}$. We call it projected because this term anticipates the structure that the model will have if it decides to eliminate a dimension by setting the current $w_{\mathrm{co}_1}^{(\text{last})}$ to zero.\\
The elimination of a latent dimension thus consists in a trade-off between the first two terms of the loss function, reflecting the reconstruction error with and without the current last latent dimension, guided by the regularization term.

\paragraph{L1 regularization: $\mathcal{L}_{\mathrm{reg}}$, $\alpha$}
For the model to understand that it has to reduce $l_\mathrm{eff}$, we have to attribute a \textit{cost}
to each dimension. This is done through the L1 regularization loss 
\begin{equation}
     \mathcal{L}_{\mathrm{reg}}(w_{\mathrm{co}}^{(\text{last})}, \alpha) = \left\| w_{\mathrm{co}}^{(\text{last})} + \alpha \right\|_1.
\end{equation}
With this regularization loss, the model is guided to minimize the L1 norm of $w_{\mathrm{co}_1}^{(\text{last})}$. Note that we add a small $\alpha \ll 1$, which pushes the model
to attribute to the unused weights the value $- \alpha$. Since Co$_1$ weights are passed through ReLU activation, we ensure that the gradient and the value are nullified after the weight
is set to $- \alpha$.

\paragraph{Orthogonality penalty:
$\mathcal{L}_{\mathrm{orth}}(f_{\theta}(x))$}
This loss is related to latent disentanglement \cite{cha2022disentanglingautoencodersdae}. Its objective is to render the latent space features to become uncorrelated of one another.
This leads to learning more generalizable representations of the data and provides a more interpretable set of features, crucial for analysis.\\
Given a latent representation matrix $f_\theta(x) \in \mathbb{R}^{n \times l}$, where $n$ is the number of samples and $l$ is the latent dimension, we define the orthogonal loss as
\begin{equation}
    \mathcal{L}_{\mathrm{orth}}(f_\theta(x)) = \left\| C_{f_\theta(x)} - I \right\|_F^2,
\end{equation}
where $I \in \mathcal{M}_{l \times l}$ is the identity matrix, $C_{f_\theta(x)} \in \mathcal{M}_{l \times l}$ is the correlation matrix of the latent representation $f_\theta(x)$ and  $\|\cdot\|_F^2$ denotes the squared Frobenius norm.

Note that the loss function (\ref{eq:loss_total}) is evaluated for each batch of input, after which its gradient is computed and the optimizer performs a step towards the minimum of (\ref{eq:loss_total}).
We will evaluate that model in the next parts in a variety of datasets with various intrinsic dimensions and relationships between features.

To assess the reconstruction accuracy of our model, we define the test loss function, which is applied to a held-out subset of the data not used during training. 
The test loss is given by
\begin{equation}
\mathcal{L}_{\mathrm{test}} = \mathcal{L}_{\mathrm{rec}}(\hat{x}, x),
\end{equation}
where $\mathcal{L}_{\mathrm{rec}}$, $\hat{x}$, and $x$ were defined in the context of the training loss function above.

Finally, Table \ref{tab:hyperparams} states the hyperparameters chosen for our model. 
\begin{table}[htp!]
    \centering
    \caption{Hyperparameters selected for the model.}
    \label{tab:hyperparams}
    \begin{tabular}{l c}
        \toprule
        \textbf{Hyperparameter} & \textbf{Value} \\
        \midrule
        Overall Learning rate & $1 \times 10^{-4}$ \\
        Learning rate of Co$_1$ & $2 \times 10^{-4}$ \\
        Batch size & 256 \\
        Number of epochs & Variable \\
        $l$ (initial dimension of the bottleneck) & Variable \\
        Optimizer & Adam \\
        $\lambda_\mathrm{rec}$ & 1 \\
        $\lambda_\mathrm{reg}$ & 0.001 \\
        $\lambda_\mathrm{orth}$ & 0.001 \\
        \bottomrule
    \end{tabular}
\end{table}
We chose a set of hyperparameters with the intention to keep them constant for all applications of our model, because one of our objectives is the versatility and adaptability of the model on different structures. 
We will only change the sample size when necessary to allow the model to learn particularly difficult structures within a time unnecessary for simpler datasets. In the following sections, we denote by $\tilde{d}$ the predicted value of $d$, the intrinsic dimension of the dataset. By definition, $\tilde{d}$ corresponds to the number of non-zero weights in Co$_1$, denoted as $l_{\mathrm{eff}}$, after the completion of training.

\section{Synthetic data tests: Legendre velocity profiles}
\label{sec:03}
In this section we evaluate the model on synthetic data. This serves as the first step towards evaluating the model on numerical solutions of vertically resolved one-dimensional free-surface flow, and compare it to the projection of the
solution on the scaled Legendre polynomial basis \cite{inproceedings}. The synthetic data in this section is thus generated by a random linear combination of scaled Legendre polynomials.

\subsection{Dataset generation}
We generated 3 different datasets with different number of intrinsic dimension ($d$) as follows:\\
Let \( \zeta \in [0, 1] \) be discretized over a grid of size $N_\zeta = 100$, and let \( P_k(\zeta) \) denote the \(k\)-th scaled Legendre polynomial, defined in \cite{TorrilhonKowalski}.\\
The dataset \( \{ f^{(j)}(\zeta) \}_{j=1}^n \) is constructed using
\begin{equation}
    f^{(j)}(\zeta) = P_1(\zeta) + \sum_{k \in \mathcal{S}} \alpha_k^{(j)} P_k(\zeta),
    \label{eq:synthpol}
\end{equation}
where \( \alpha_k^{(j)} \sim \mathcal{U}(0, 1) \) are i.i.d. uniform random variables for each sample \( j \in \{1, \ldots, n\} \), $n$ being the number of samples.\\
The set of polynomials $\mathcal{S} \in\mathcal{P}(\mathbb{N})$ is predefined for each of the three datasets generated. As each random variable $\alpha_k$ is independent from the others, each one contributes to one independent dimension, thus we have the intrinsic dimension $d = |\mathcal{S}|$.\\
We generate datasets of size \( n = 20{,}000 \) of the intrinsic dimensions given in Table \ref{tab:model_performance}.

\begin{table}[htbp!]
    \centering
    \caption{Summary of synthetic dataset characteristics and IDEA reconstruction errors (mean squared error).}
    \label{tab:model_performance}
    \small
    \setlength{\tabcolsep}{5pt} 
    \begin{tabular}{ccccccc}
        \toprule
        \textbf{\#} & \textbf{$\mathcal{S}$} & \textbf{$d$} & \textbf{$\tilde d$} & \multicolumn{3}{c}{$\mathcal{L}_{\text{test}}$} \\
        \cmidrule(lr){5-7}
                   &                        &              &                     &    $\tilde d{+}1$ & $\tilde d$ & $\tilde d{-}1$ \\
        \midrule
         1 & $\{3\}$ & 1 & 1 &$7.3\!\cdot\!10^{-7}$ & $7.8\!\cdot\!10^{-7}$ & $1.28\!\cdot\!10^{-2}$ \\
         2 & $\{3, 5\}$ & 2 & 2 & $1.3\!\cdot\!10^{-6}$ & $1.5\!\cdot\!10^{-6}$ & $2.0\!\cdot\!10^{-3}$ \\
         3 & $\{3, 5, 6, 7\}$ & 4 & 4 & $9.6\!\cdot\!10^{-6}$ & $9.7\!\cdot\!10^{-6}$ & $1.3\!\cdot\!10^{-3}$ \\
        \bottomrule
    \end{tabular}
\end{table}

All instances of the models were initialized with a latent size $l = 8$. For these three datasets, IDEA successfully predicted the correct intrinsic dimension $\tilde d=d$ as shown in Table \ref{tab:model_performance}. Note that since the polynomials are evaluated on a 100-point grid, the embedded dimension is $p =100$ here.

\subsection{Model performance}
A distinctive feature of our model is that it does not only successfully predict the intrinsic dimension, but it also reconstructs the data.
In fact, when the model estimates a specific dimension $\tilde d$, the training loop outputs the best model with $\tilde d-1$, $\tilde d$ and $\tilde d+1$. For the synthetic data test case, this leads to the results shown in Table \ref{tab:model_performance}.

The results in table (\ref{tab:model_performance}) show that the predicted intrinsic dimension $\tilde d$ is indeed correct, as a reconstruction with $\tilde d - 1$ intrinsic variables leads to substantial reconstruction loss, while a reconstruction with $\tilde d + 1$ intrinsic dimensions does not lead to an additional gain in accuracy.

\begin{figure}[htbp!]
\centering
\begin{tikzpicture}
  \begin{axis}[
    width=0.9\linewidth,
    height=6cm,
    xlabel={Epoch},
    ylabel={$\mathcal{L}_{\text{total}}$},
    grid=both,
    thick,
    ymode=log,
    enlargelimits=0.05,
    legend pos=north east
  ]
    \addplot[
      blue,
      mark=*,
      mark size=1pt,
      only marks,
    ] table [x=Epoch, y=Loss, col sep=comma] {body/figures/Loss_curve.csv};
  \end{axis}
\end{tikzpicture}
\caption{Synthetic dataset 3 total train loss $\mathcal{L}_{\text{total}}$ as a function of training epoch.}
\label{fig:loss_curve}
\end{figure}
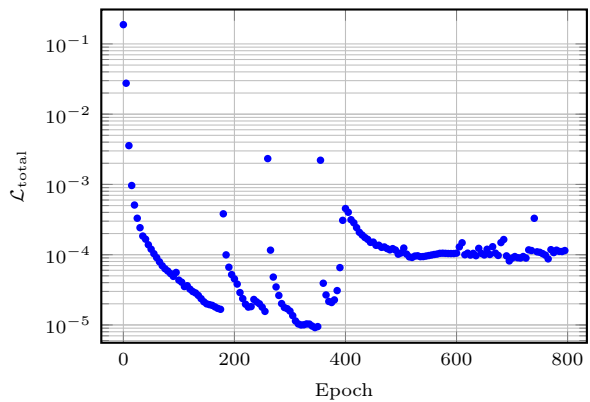

Fig. \ref{fig:loss_curve} shows the evolution of the test loss $\mathcal{L}_{\text{test}}$ during the training on dataset 3. We obtain from the plot that the training can be characterized by two distinct phases:\\
\paragraph{Phase 1: Epoch 1-500}
In the first phase, the network progressively suppresses non-contributing latent dimensions. Each time one latent variable is eliminated ($l_{\text{eff}}$ decreases by one), the loss function spikes because the network adjusts the modified loss function.
\paragraph{Phase 2: Epoch 500-800}
In the last phase, the network stops eliminating dimensions because it no longer considers the trade-off between reconstruction loss and projected reconstruction loss to be interesting.
We can see that it is attempting to project the latent space onto 3 dimensions, which appears to be insufficient for accurate reconstruction. The test loss oscillates around its local minimum.

\subsection{Latent space disentanglement}
We added a term for latent disentanglement $\mathcal{L}_{\text{orth}}$ in the training loss function \cref{eq:loss_total}, promoting zero Pearson correlation among latent variables. 
After training, by computing the latent representation of the data ($f_{\theta}(x)$), we can plot a correlation matrix of the latent representation. 
We will also observe the correlation between the latent variables $L_i$, for $i=0,\ldots, \tilde d-1$ learned by the model and the original coefficient $\alpha_i$, for $i\in S$ in front of each polynomial (generated by uniform distribution) from \cref{eq:synthpol}.

\begin{figure}[htbp!]
\centering
\definecolor{mygray}{HTML}{B7B2B2}

\begin{tikzpicture}[scale=0.6]

    \shade[top color=myviolet,bottom color=lightgray]
        (9.2,-8.5) rectangle (9.7,-0.5);
    \draw[black] (9.2,-8.5) rectangle (9.7,-0.5);
    
    \node[anchor=west] at (9.7,-0.5) {1};
    \node[anchor=west] at (9.7,-8.5) {0};

  \foreach \y [count=\n] in {
{1.0, 0.0,  0.0, -0.0, -0.3,  0.5, -0.6, -0.1},
{ 0.0,  1.0, -0.0,  0.0, -0.8, -0.4, -0.0, -0.3},
{ 0.0, -0.0,  1.0,  0.0,  0.2, -0.4, -0.5,  0.3},
{-0.0,  0.0,  0.0,  1.0, -0.0, -0.0, -0.0, -0.0},
{-0.3, -0.8,  0.2, -0.0,  1.0, -0.0, -0.0,  0.0},
{ 0.5, -0.4, -0.4, -0.0, -0.0,  1.0, -0.0, -0.0},
{-0.6, -0.0, -0.5, -0.0, -0.0, -0.0,  1.0, -0.0},
{-0.1, -0.3,  0.3, -0.0,  0.0, -0.0, -0.0,  1.0},
    } {

      \foreach \x [count=\m] in \y {

        \pgfmathsetmacro{\val}{abs(\x)*100}
        \node[
          fill=myviolet!\val!lightgray, 
          minimum width=6mm, minimum height=6mm,
          text width=6mm,
          align=center,
          text=White,
          font=\footnotesize,
          inner sep=0pt,
          outer sep=0pt,
          draw=gray
        ] at (\m,-\n) {\x};
      }
    }

  \foreach \a [count=\i] in {$L_0$,$L_1$,$L_2$,$L_3$,$\alpha_3$,$\alpha_5$,$\alpha_6$,$\alpha_7$} {
    \node[minimum size=6mm] at (0,-\i) {\a};
  }

  \foreach \label [count=\i from 1] in {$L_0$,$L_1$,$L_2$,$L_3$,$\alpha_3$,$\alpha_5$,$\alpha_6$,$\alpha_7$} {
  \node[minimum size=6mm] at (\i, 0) {\label};
}

\end{tikzpicture}
\caption{Correlation matrix of dataset 3, between the four latent variables $L_i$ and the projection coefficients on the scaled Legendre polynomial basis $\alpha_i$ (\cref{eq:synthpol}). The color varies according to the absolute value of the correlation between the parameters.}
\label{fig:corr1}
\end{figure}
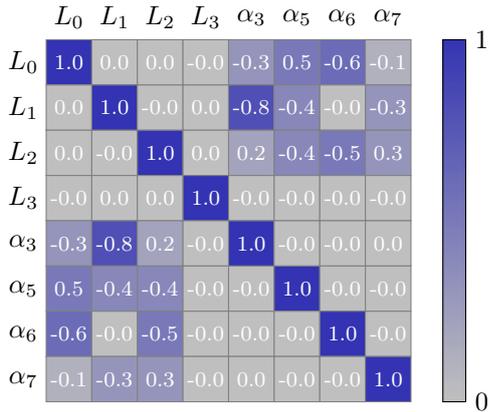

We can see in Fig. \ref{fig:corr1} that even though the latent space is completely uncorrelated and so are the $\alpha_i$'s, 
the model learns a combination of coefficients, making it difficult to link the latent variables of the model to the coefficients of the orthogonal projection. We note that while $L_3$ is almost uncorrelated with the $\alpha_i$, it still encodes information about them, as shown in \cref{tab:model_performance}, where the reconstruction without $L_3$ yields a much higher test loss. We recall that Pearson correlation captures only linear relationships; thus, a correlation of zero does not imply independence, but may instead indicate the presence of nonlinear dependencies.

After the successful synthetic data tests, we continue with benchmark tests. 

\section{Benchmark tests: structured manifolds}
\label{sec:4}
\subsection{Choice of dataset and benchmark comparison}

We perform a comparative evaluation of our model with the benchmark framework applied to relevant state-of-the-art intrinsic dimension estimators presented in \cite{article}.
The model is evaluated on synthetic datasets provided by the module \texttt{scikit-dimension} presented in \cite{Bac_2021} which reuses those from \cite{article}.
The synthetic datasets are generated by drawing samples from manifolds ($\mathcal{M}$) linearly or nonlinearly embedded in higher-dimensional spaces, see \cref{tab:bm1}. The original paper did not evaluate the estimators on $\mathcal{M}_{5a}$. For this dataset, we reproduced the estimation using the same parameters to report the results. \textbf{\texttt{TwoNN}} and  \textbf{\texttt{lPCA}} are not reported in \cite{article}, we used the default hyperparameters from \texttt{scikit-dimension}.

As shown in \cref{tab:bm1}, IDEA accurately predicts the intrinsic dimension of the datasets. While baseline estimators can have blind spots on specific datasets, leading to significant estimation errors, IDEA consistently approximates the true intrinsic dimension on all of these datasets.
We note that for complex nonlinear manifolds, a larger number of samples was sometimes required to achieve accurate prediction. This led to increased computation time (see \ref{sec:BM}, where we illustrate how prediction accuracy evolves with sample size).

\subsection{Comparison against standard deep learning based methods}

Aside from the evaluation with the scikit-dimension models, we trained other deep-learning-based intrinsic dimension estimators and reported the results, namely: Variational autoencoder (VAE), Sequential autoencoder (SAE).
\paragraph{Variational autoencoder (VAE)} 
Presented in \cite{Diederik_2019}, Variational autoencoder is an autoencoder with a probabilistic latent space. Each input is mapped via the Encoder to $\mu, \sigma$, then we sample from a distribution $\mathcal{N}(\mu, \sigma^2)$ and the sampled latent vector is passed through the decoder. Initially, VAE were invented to generate new samples directly by sampling from the latent space. The probabilistic aspect of the latent space makes it more structured. For our analysis, we will estimate the intrinsic dimension by counting the number of latent dimensions for which the mean estimated standard deviation is above a 0.05 threshold. Hyperparameters are: batch size $=256$, learning rate $=10^{-3}$, number of epochs $=300$, hidden width $=32$ (two hidden layers in both encoder and decoder), activation LeakyReLU$(0.2)$. The regularization term is the standard KL divergence to $\mathcal{N}(0,I)$.

\paragraph{Sequential autoencoder (SAE)}

For the SAE baseline, we train a standard autoencoder for latent sizes $l\in\{1,\dots,p\}$, similar to \cite{koellermeier2023modelorderreduction1d} and select the smallest $l$ such that the reconstruction gain $L_{0}-L_{l}$ reaches at least $95\%$ of the maximum gain $L_{0}-L_{p}$. The autoencoder uses the same hyperparameters as IDEA and is trained for 400 epochs for each latent sizes on $5\times 10^{4}$ samples.

\renewcommand{\arraystretch}{2} 

\begin{table*}[t]
    \centering
    \small 
    \caption{Benchmark datasets characteristics (intrinsic dimension $d$ and embedded dimension $p$), and result comparison of the intrinsic dimension estimation $\tilde{d}$ between IDEA and state-of-the-art from \cite{article} including new benchmark dataset of embedded helix $(\mathcal{M}_{5a})$. Green cells denote an exact estimation (within $d \pm 0.05$), whereas cells in red indicate an error exceeding one dimension from the true value.}
    \label{tab:bm1}
    \rowcolors{2}{white}{gray!5}
    \setlength{\tabcolsep}{6pt} 

    \resizebox{0.80\textwidth}{!}{
\begin{tabular}{l m{7.5cm} c c c c c c c c c c c}
\toprule
\textbf{Name} & \textbf{Description} & \textbf{$d$} & \textbf{$p$}
& \multicolumn{3}{c}{\textbf{Deep learning methods}}
& \multicolumn{6}{c}{\textbf{Probabilistic / classical methods}} \\
\cmidrule(lr){5-7}\cmidrule(lr){8-13}
& & & 
& \cellcolor{CornflowerBlue!20}\textbf{IDEA} & \textbf{VAE} & \textbf{SAE}
& \textbf{\texttt{MLE}} & \textbf{\texttt{TwoNN}} & \textbf{\texttt{CD}} & \textbf{\texttt{MIND$_{KL}$}} & \textbf{\texttt{DANCo}} & \textbf{\texttt{lPCA}} \\
\midrule

$\mathcal{M}_1$      & 10D sphere linearly embedded
& 10 & 11 & \cellcolor{green!20}$10.0$ & 9.00 & \cellcolor{green!20}$10.0$
& $9.10$ & \cellcolor{green!20}$9.97$ & $9.12$ & $10.30$ & $10.09$ & $11.00$ \\

$\mathcal{M}_2$      & Affine space
& 3  & 5  & \cellcolor{green!20}$3.0$  & \cellcolor{green!20}$3.00$ & \cellcolor{green!20}$3.0$
& $2.88$ & \cellcolor{green!20}$2.98$ & $2.88$ & \cellcolor{green!20}$3.00$ & \cellcolor{green!20}$3.00$ & \cellcolor{green!20}$3.00$ \\

$\mathcal{M}_3$      & Concentrated figure, mistakable with a 3D one
& 4  & 6  & $5.0$ & \cellcolor{green!20}$4.00$ & \cellcolor{green!20}$4.00$
& $3.83$ & $3.77$ & $3.23$ & \cellcolor{green!20}$4.00$ & \cellcolor{green!20}$4.00$ & $5.00$ \\

$\mathcal{M}_4$      & Nonlinear manifold
& 4  & 8  & \cellcolor{green!20}$4.0$ & \cellcolor{red!20}$7.00$ & 4.8
& \cellcolor{green!20}$3.95$ & $3.80$ & $3.88$ & $4.15$ & \cellcolor{green!20}$4.00$ & \cellcolor{red!20}$8.00$ \\

$\mathcal{M}_{5b}$   & 2D helical surface embedded in 3D
& 2  & 3  & \cellcolor{green!20}$2.0$ & \cellcolor{green!20}$2.00$ & \cellcolor{green!20}$2.0$
& \cellcolor{green!20}$1.97$ & \cellcolor{green!20}$2.02$ & \cellcolor{green!20}$1.98$ & \cellcolor{green!20}$2.00$ & \cellcolor{green!20}$2.00$ & $3.00$ \\

$\mathcal{M}_6$      & Nonlinear manifold
& 6  & 36 & \cellcolor{green!20}$6.0$ & \cellcolor{red!20}$12.00$ & \cellcolor{red!20}$9.00$
& $6.39$ & $5.86$ & $5.91$ & $6.50$ & $7.00$ & \cellcolor{red!20}$12.00$ \\

$\mathcal{M}_7$      & Swiss Roll
& 2  & 3  & \cellcolor{green!20}$2.0$ &  \cellcolor{green!20}$2.0$ & \cellcolor{green!20}$2.0$
& \cellcolor{green!20}$1.96$ & \cellcolor{green!20}$2.00$ & $1.93$ & $2.07$ & \cellcolor{green!20}$2.00$ & $3.00$ \\

$\mathcal{M}_{Beta}$ & sample of $N$ points in $[0, 1)^{10}$, according to a beta distribution $\beta_{0.5,10}$
& 10 & 40 & \cellcolor{green!20}$10.0$ & 9.67 & \cellcolor{red!20}$7.80$
& \cellcolor{red!20}$6.36$ & \cellcolor{red!20}$6.77$ & \cellcolor{red!20}$3.05$ & \cellcolor{red!20}$7.00$ & \cellcolor{red!20}$7.00$ & \cellcolor{green!20}$10.00$ \\

$\mathcal{M}_{P3}$   & 3D paraboloid, nonlinearly embedded according to a multivariate Burr distribution ($\alpha=1$)
& 3  & 12 & \cellcolor{green!20}$3.0$ & \cellcolor{green!20}$3.00$ & \cellcolor{green!20}$3.00$
& $2.89$ & \cellcolor{green!20}$2.98$ & $2.43$ & \cellcolor{green!20}$3.00$ & \cellcolor{green!20}$3.00$ & \cellcolor{red!20}$1.00$ \\

$\mathcal{M}_{P6}$   & 6D paraboloid, nonlinearly embedded according to a multivariate Burr distribution ($\alpha=1$)
& 6  & 21 & \cellcolor{green!20}$6.0$ & \cellcolor{green!20}$6.00$ & \cellcolor{green!20}$6.00$
& \cellcolor{red!20}$4.96$ & $5.44$ & \cellcolor{red!20}$3.58$ & $5.04$ & \cellcolor{green!20}$6.00$ & \cellcolor{red!20}$1.00$ \\

\hline\hline
$\mathcal{M}_{5a}$   & 1D helix embedded in 3D
& 1  & 3  & $2.0$ & $2.0$ & $2.00$
& $1.06$ & \cellcolor{green!20}$0.99$ & \cellcolor{green!20}$1.00$ & \cellcolor{green!20}$1.00$ & \cellcolor{green!20}$1.00$ & \cellcolor{red!20}$3.00$ \\

\bottomrule
\end{tabular}
}
\end{table*}

\renewcommand{\arraystretch}{1}

As our model is able to reconstruct the dataset under the estimated dimension, we can also compute reconstruction losses, see \ref{sec:BM}.

\begin{figure}[htbp!]
    \centering
    \scalebox{0.8}{
    \begin{subfigure}[h]{\picWidth}
    
    \includegraphics[width=1\linewidth]{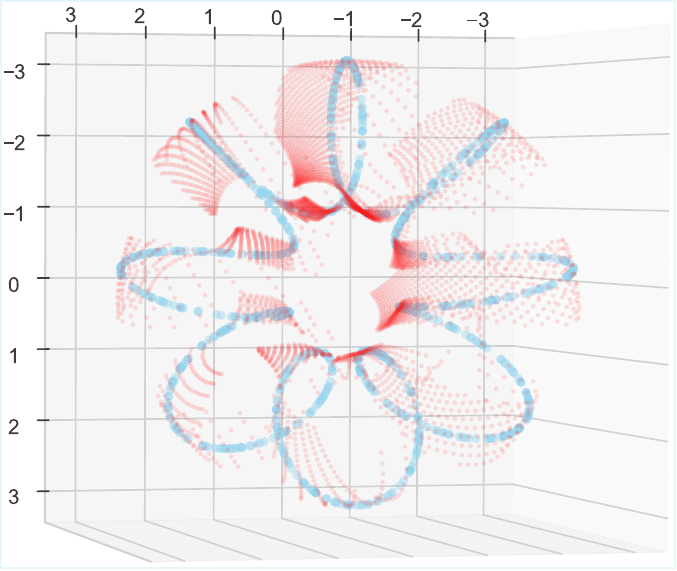}
    \caption{Representation of the one-dimensional helix dataset (sky blue) and model's reconstruction of the complete latent space with $l_{\text{eff}}=2$ (red).}
    \end{subfigure}}
    

    \scalebox{0.8}{
    \begin{subfigure}[h]{\picWidth}
    \includegraphics[width=1\linewidth]{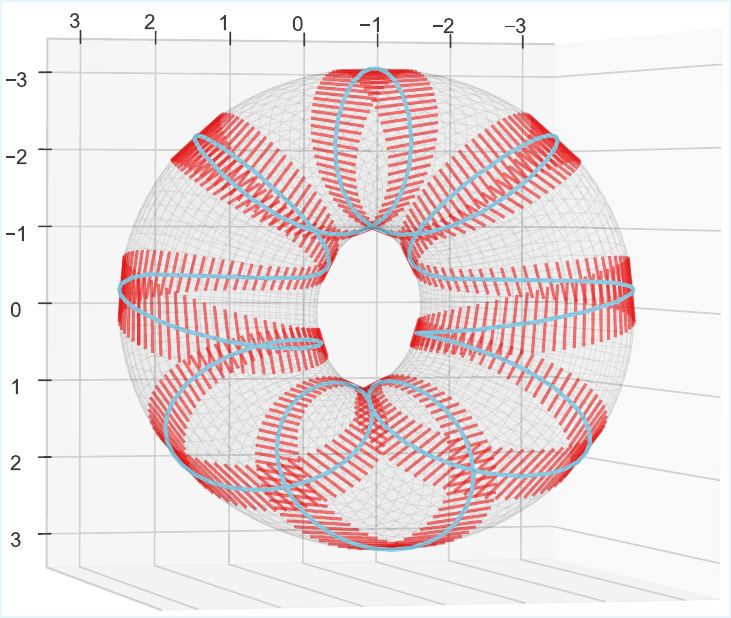}
    \caption{Theoretical representation of the input dataset (sky blue), the manifold in which IDEA projected the data (red), and the subspace of the known two-dimensional manifold torus (light grey).}
    \end{subfigure}}
    
    \caption{Representations of the estimated latent space of IDEA upon the one-dimensional helix ($\mathcal{M}_{5a}$)(a) and the theoretical manifold from which it is taken (b).}
    \label{fig:tore}
\end{figure}

\subsection{Analyzing nonlinear reconstruction and possible improvements}

IDEA exhibited difficulties with the one-dimensional helix $\mathcal{M}_{5a}$, a complex nonlinear helix embedded in a 3D space.
IDEA predicted an intrinsic dimension $\tilde{d} = 2$ for this one-dimensional manifold. By evaluating the decoder on a grid of points spanning the latent space, we observed (Fig. \ref{fig:tore}) that IDEA identified a manifold $\mathcal{S_T}$ that can be mapped into $\mathbb{R}^2$ and contains the helix ($\mathcal{H}_1$). Geometrically, $\mathcal{S_T}$ resembles a ribbon surrounding the two-dimensional torus along the poloidal direction. Formally, a topological manifold of dimension d is a topological space $M$ such that every point $p \in M$ has a neighborhood $U$ homeomorphic to an open subset of $\mathbb{R}^d$. This condition holds for the torus, and here the data-oriented “intrinsic dimension” coincides with the topological dimension of the projected space, namely 2. IDEA found a manifold $\mathcal{S_T}$ of dimension less than $p$ which contains the data. However, this manifold $\mathcal{S_T}$ does not align with the neighborhood structure of $\mathcal{H}_1$ which is in fact a one-dimensional manifold.

While the performance of IDEA is very good also in comparison with the state-of-the-art, several strategies could be investigated to further improve the model's performance on special nonlinear manifolds, provided that the potential increase in computation time is acceptable:\\

First, the model exhibits significant sensitivity to the learning rate of the bottleneck layer weights, which is currently set to $2 \cdot 10^{-4}$, while the global learning rate is fixed at $1 \cdot 10^{-4}$. To better adapt to the characteristics of each dataset, we propose the integration of more advanced learning rate tuning techniques such as learning rate warm-up~\cite{unknown}, learning rate scheduling,
and cyclical learning rates~\cite{smith2017cyclicallearningratestraining}. These approaches could increase the model’s flexibility by allowing the learning rate to adapt dynamically during training, thus reducing the need for manual tuning.\\

As the bottleneck learning rate is a hyperparameter, we don't want the user to have to tune it on a given use case. Although incorporating dynamic learning rate behavior can make this hyperparameter more generalizable to handle very different datasets. We propose an ablation study on datasets $\mathcal{M}_3$ and $\mathcal{M}_{5a}$ in Table \ref{tab:ablation_sched_m3_m5a}


The learning rate is defined with two parameter groups: (i) the base network parameters, trained with a fixed learning rate $\eta_{\text{base}}$, and (ii) the CancelOut/Regularizer parameters, trained with a learning rate $\eta_{\text{reg}}(t)$. To isolate the sensitivity of the bottleneck layer, we schedule \emph{only} $\eta_{\text{reg}}(t)$ while keeping $\eta_{\text{base}}$ constant throughout training. Let $t \in \{1,\dots,T\}$ denote the scheduler step index, incremented either once per epoch or once per mini-batch depending on the strategy.

In the benchmark experiments in Table~\ref{tab:bm1}, $\eta_{\text{reg}}(t)$ was kept constant and set to $2\times 10^{-4}$. We observe that increasing $\eta_{\text{reg}}$ makes the regularizer more aggressive in pruning latent dimensions, as the associated weight updates become larger. This can improve performance on challenging datasets such as $\mathcal{M}_3$ and $\mathcal{M}_{5a}$, but may lead to over-compression on datasets that are already well estimated. A learning-rate schedule can adapt $\eta_{\text{reg}}(t)$ during training, reducing the need for manual tuning and improving the generalization across different datasets.

We tested various scheduling methods as follows. We show ablation results for challenging datasets $\mathcal{M}_3$ and $\mathcal{M}_{5a}$ in Table \ref{tab:ablation_sched_m3_m5a}.

{\setlength{\itemsep}{1pt}\setlength{\parskip}{0pt}\setlength{\parsep}{0pt}%
 \setlength{\leftmargini}{1.2em}\setlength{\labelsep}{0.4em}%
 \renewcommand{\labelitemi}{\raisebox{0.2ex}{\tiny$\bullet$}}%
\begin{itemize}
\item \textbf{Baseline:} $\eta_{\text{reg}}(t)=\eta_{\text{reg}}^{(0)}$. \;\textit{hp:} $\eta_{\text{reg}}^{(0)}=2\times10^{-4}$.
\item \textbf{Step decay:} $\eta_{\text{reg}}(t)=\eta_{\text{reg}}^{(0)}\,\gamma^{\lfloor t/S\rfloor}$. \;\textit{hp:} $\eta_{\text{reg}}^{(0)}=2\times10^{-2}$, $S=150$, $\gamma=0.1$.
\item \textbf{Warmup+Cosine:} $t\le W$: $\eta_{\text{reg}}(t)=\eta_{\text{reg}}^{\min}+(\eta_{\text{reg}}^{(0)}-\eta_{\text{reg}}^{\min})\,t/W$;\;
$t>W$: $\eta_{\text{reg}}(t)=\eta_{\text{reg}}^{\min}+\tfrac12(\eta_{\text{reg}}^{(0)}-\eta_{\text{reg}}^{\min})\!\left(1+\cos\!\bigl(\pi\frac{t-W}{T-W}\bigr)\right)$. \;\textit{hp:} $\eta_{\text{reg}}^{\min}=2\times10^{-6}$, $\eta_{\text{reg}}^{(0)}=2\times10^{-2}$, $W=0.1T$.
\item \textbf{OneCycle:} let $\eta_{\text{reg}}^{\mathrm{init}}=\eta_{\text{reg}}^{\max}/\texttt{div}$ and $\eta_{\text{reg}}^{\mathrm{final}}=\eta_{\text{reg}}^{\mathrm{init}}/\texttt{final\_div}$. 
$T_\uparrow=\lfloor pT\rfloor$;\; $t\le T_\uparrow$: $\eta_{\text{reg}}(t)=\eta_{\text{reg}}^{\mathrm{init}}+(\eta_{\text{reg}}^{\max}-\eta_{\text{reg}}^{\mathrm{init}})\,t/T_\uparrow$;\;
$t>T_\uparrow$: $\eta_{\text{reg}}(t)=\eta_{\text{reg}}^{\mathrm{final}}+\tfrac12(\eta_{\text{reg}}^{\max}-\eta_{\text{reg}}^{\mathrm{final}})\!\left(1+\cos\!\bigl(\pi\frac{t-T_\uparrow}{T-T_\uparrow}\bigr)\right)$.
\;\textit{hp:} $\eta_{\text{reg}}^{\max}=2\times10^{-1}$, $p=0.1$, \texttt{div}=$10^3$, \texttt{final\_div}=$1$.
\end{itemize}}


\begin{table}[h]
\centering
\small
\setlength{\tabcolsep}{7pt}
\caption{Ablation study of learning-rate schedules applied \emph{only} to the CancelOut/Regularizer parameter group. We report the estimated intrinsic dimension $\hat d$ (mean $\pm$ std over 5 repeated runs/seeds) for the two challenging datasets $\mathcal{M}_3$ and $\mathcal{M}_{5a}$ (noise level: 0).}
\label{tab:ablation_sched_m3_m5a}
\begin{tabular}{lcc}
\toprule
\textbf{Scheduler} & \textbf{$\mathcal{M}_3$} ($d=4$) & \textbf{$\mathcal{M}_{5a}$} ($d=1$) \\
\midrule
Baseline            & $5.00 \pm 0 $ & $2.00 \pm 0 $ \\
StepLR              & $4.50 \pm 0.71 $ & $2.00 \pm 0$ \\
Warmup+Cosine       & $4.86 \pm 0.38 $ & $2.00 \pm 0$ \\
OneCycle            & $3.07 \pm 1.49 $ & $1.54 \pm 0.52$ \\
\bottomrule
\end{tabular}
\end{table}

These results show that the learning-rate schedule applied to the Regularizer has a clear impact on IDEA’s ID estimate: compared to the Baseline, StepLR and Warmup+Cosine shift the mean prediction closer to the ground truth on $\mathcal{M}_3$. More aggressive schedules such as OneCycle, while increasing accuracy on $\mathcal{M}_{5a}$ substantially increases the run-to-run variability (larger std), indicating reduced robustness and less stable dimension selection.
\section{Application test: vertically resolved free-surface flow}
\label{sec:5}
\subsection{Reference system and data generation}
Shallow flow models are widely used in applications such as weather forecasting and simulation-based natural hazard assessment \cite{TorrilhonKowalski}.
Our application case considers a vertically resolved one-dimensional free-surface shallow flow PDE model, which describes the evolution of the vertical velocity profile \( u(t, x, \zeta) \) and the free-surface height \( h(t, x) \) over time $t$ and along the $x$ axis, where \( \zeta \in [0,1] \) denotes a normalized vertical coordinate.\\
According to \cite{TorrilhonKowalski}, The governing equations are
\begin{align}
\partial_t h + \partial_x (h u) + \partial_\zeta (h \omega) &= 0,  \\
\partial_t (h u) + \partial_x \left( h u^2 + \frac{1}{2} h^2 \right) + \partial_\zeta (h u \omega) &= \frac{R}{h} \, \partial_{\zeta\zeta} u,
\end{align}
subject to the boundary conditions 
\begin{equation}
\left. \partial_\zeta u \right|_{\zeta = 0} = \frac{h}{\chi} u|_{\zeta = 0}, \quad
\left. \partial_\zeta u \right|_{\zeta = 1} = 0.
\end{equation}
The initial condition of a smooth wave is given by: 
\begin{equation}
h(x, 0) = 1 + \exp\left(3 \cos\left(\pi(x + x_0)\right) - 4\right),
\end{equation}
where $\omega(t, x, \zeta)$ is a vertical velocity coupling term, $R$ is the friction coefficient and $\chi$ is the slip length. We refer to \cite{TorrilhonKowalski} for more details.

While the reference system from \cite{TorrilhonKowalski} can be discretized pointwise in the \(\zeta\) direction, often only a few macroscopic variables can describe the full vertical velocity profile, in which case their number corresponds to the ID of the profiles. One example is the vertical average \(u_m(t, x)\), which could be computed purely from the reduced, so-called shallow water equations. 
However, such simple models often fail to capture important vertical dynamics in realistic settings \cite{inproceedings}.

More accurate models, such as the moment approximation presented in \cite{TorrilhonKowalski}, project the reference solution onto a subset $\mathcal{S}$ of the scaled Legendre polynomial basis, yielding coordinates (also called moments) $\alpha_i$ for each $i \in \mathcal{S}$. However, such models may use more parameters than the intrinsic dimension of the profile, which is expected to lie on a 2-dimensional manifold (along $x$ and $t$).

To address this limitation, we propose using IDEA to learn a compact and informative latent representation of the vertical velocity profile. 
Specifically, we apply our model to data generated from a pointwise numerical discretization of the vertically resolved system, on a grid of size \( n_t \times n_x \times n_\zeta \), as in \cite{TorrilhonKowalski}. 
The resulting solution is transformed into a dataset represented as a matrix \( M \in \mathcal{M}_{n_x \cdot n_t, n_\zeta + 1}(\mathbb{R}) \). The first column of the dataset is the water height $h(x_i, t_i)$ evaluated for each point of the grid along $x$ and $t$.
IDEA successfully balances computational efficiency and reconstruction accuracy by learning a precise mapping of the full solution into a lower-dimensional latent space of size \(1 < d < n_{\zeta}\), thus identifying the ID of the velocity profiles.

\subsection{Model specifications for application to free-surface flows}

The objective is to reduce the dimensionality of the velocity profile space, using our intrinsic estimator model IDEA. As each point has one water height feature, we explicitly reserve one latent dimension for representing the water height, and encourage the model to learn this value by adding a term in the loss function ($\mathcal{L}_h$) which penalizes how far the first latent variable ($L_0$, first coordinate of the output vector of the layer $\text{Co}_2$) deviates from $h(x, t)$ using the MSE.

The loss function therefore reads
\begin{equation}
    \mathcal{L}_{\text{SWE}} = \underbrace{\mathcal{L}_{\text{total}}(\hat{x}, x,\hat{x}_{\text{adj}}, w_{\text{co}_1}^{(\text{last})})}_{\text{initial loss}} + 
    \underbrace{ \mathcal{L}_{h}(L_0,h(x,t))}_{\text{Water height Loss}},
\end{equation}
where $\mathcal{L}_{\text{total}}$ is the loss previously defined, and $\mathcal{L}_{h}$ is the additional loss for the deviation of the first latent variable from the water height.

\subsection{Results and reconstruction analysis}
We consider two cases.

\subsubsection{First case: simple solution with a small number of moments ($\mathcal{M}_1$)}
\begin{figure}[htp!]
	\setlength\figureheight{0.2\textwidth}
	\setlength\figurewidth{0.2\textwidth}
	\centering
    \scalebox{0.75}{\begin{tikzpicture}
  \begin{axis}[
    width=\linewidth,
    height=6cm,
    xlabel={t},
    ylabel={$\alpha_i$},
    legend style={at={(1.05,1)}, anchor=north west},
    grid=both,
    thick,
    cycle list name=color list
  ]
    \addplot table [x=time, y ={Alpha 0}, col sep=comma] {body/figures/moments_1.csv};
    \addlegendentry{$\alpha_0$}
    
    \addplot table [x=time, y={Alpha 1}, col sep=comma] {body/figures/moments_1.csv};
    \addlegendentry{$\alpha_1$}
    
    \addplot table [x=time, y={Alpha 2}, col sep=comma] {body/figures/moments_1.csv};
    \addlegendentry{$\alpha_2$}
    
    \addplot table [x=time, y={Alpha 3}, col sep=comma] {body/figures/moments_1.csv};
    \addlegendentry{$\alpha_3$}
    
    \addplot table [x=time, y={Alpha 4}, col sep=comma] {body/figures/moments_1.csv};
    \addlegendentry{$\alpha_4$}

    \addplot table [x=time, y={Alpha 5}, col sep=comma] {body/figures/moments_1.csv};
    \addlegendentry{$\alpha_5$}

  \end{axis}
\end{tikzpicture}}
    \caption{Free-surface flow: data evolution with time of the 6 first projection coefficient $\alpha_i$, averaged along x}
    \label{fig:frt}
\end{figure}
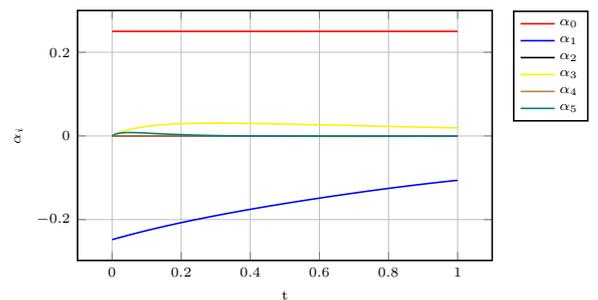

IDEA is first evaluated on a solution that can be accurately reconstructed using only a few moments. By setting a large slip length $\chi = 10^6$, we eliminate bottom friction entirely, allowing the profile to evolve more freely and preserve a structure that stabilizes quickly.
Fig. \ref{fig:frt} shows that from time $t\approx 0.3$, the only significant moments are $\alpha_0$, $\alpha_1$ and $\alpha_3$. IDEA is then applied to the data generated by the reference solution for $t \in \left[ 0.3, 1 \right] $, starting with $l_\text{eff}=8$. The corresponding results are presented in Fig. \ref{tab:tl3}.

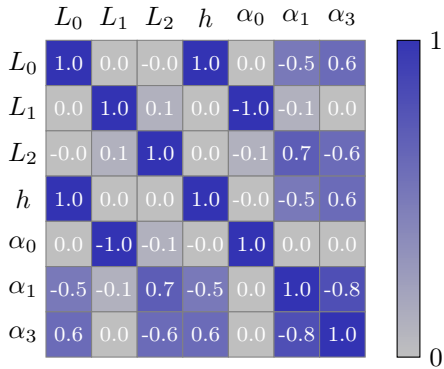
\begin{figure}[htp!]
\centering
\begin{tikzpicture}[scale=0.6]

  \shade[top color=myviolet,bottom color=lightgray]
      (8.2,-7.5) rectangle (8.7,-0.5);
  \draw[black] (8.2,-7.5) rectangle (8.7,-0.5);

  \node[anchor=west] at (8.7,-0.5) {1};
  \node[anchor=west] at (8.7,-7.5) {0};

  \foreach \y [count=\n] in {
{ 1.0 ,  0.0 , -0.0,  1.0 ,  0.0 , -0.5,  0.6},
{ 0.0 ,  1.0 ,  0.1,  0.0, -1.0, -0.1,  0.0},
{-0.0,  0.1,  1.0 ,  0.0, -0.1,  0.7, -0.6},
{ 1.0 ,  0.0,  0.0,  1.0 , -0.0 , -0.5,  0.6},
{ 0.0 , -1.0, -0.1, -0.0 ,  1.0 ,  0.0 ,  0.0 },
{-0.5, -0.1,  0.7, -0.5,  0.0 ,  1.0 , -0.8},
{ 0.6,  0.0, -0.6,  0.6,  0.0 , -0.8,  1.0 },
    } {

      \foreach \x [count=\m] in \y {

        \pgfmathsetmacro{\val}{abs(\x)*100}
        \node[
          fill=myviolet!\val!lightgray, 
          minimum width=6mm, minimum height=6mm,
          text width=6mm,
          align=center,
          text=White,
          font=\footnotesize,
          inner sep=0pt,
          outer sep=0pt,
          draw=gray
        ] at (\m,-\n) {\x};
      }
    }

  \foreach \a [count=\i] in {$L_0$,$L_1$,$L_2$,$h$,$\alpha_0$,$\alpha_1$,$\alpha_3$} {
    \node[minimum size=6mm] at (0,-\i) {\a};
  }

  \foreach \label [count=\i from 1] in {$L_0$,$L_1$,$L_2$,$h$,$\alpha_0$,$\alpha_1$,$\alpha_3$} {
  \node[minimum size=6mm] at (\i, 0) {\label};
}
\end{tikzpicture}
\caption{Pearson correlation matrix between the three latent variables $L_i$ and the coefficients $a_i$. The color varies according to the absolute value of the correlation between the parameters.}
\label{fig:corr2}
\end{figure}

\begin{table*}[htp!]
    \centering
    \caption{Summary of reconstruction error across application test datasets and the projected dimensions.}
    \label{tab:tl3}
    \small
    \setlength{\tabcolsep}{6pt}
    \renewcommand{\arraystretch}{1.2}
    \begin{tabular}{l c c c c c c c c c}
        \toprule
        \textbf{Dataset} 
        & \textbf{$d$} 
        & \textbf{$\tilde{d}$} 
        & \textbf{$l$}
        & \textbf{$t_0$}
        & \textbf{\# Epochs}
        & \textbf{\# Samples}
        & \multicolumn{3}{c}{\textbf{$\mathcal{L}_{\text{test}}$}} \\
        \cmidrule(lr){8-10}
        & & & & & & & $\tilde d+1$ & $\tilde d$ & $\tilde d-1$ \\
        \midrule
        $\mathcal{M}_1$ & 3 & 3 & 8 & 0.3 & 1000 & 24600& $3.50 \times 10^{-6}$ & $6.36 \times 10^{-6}$ & $4.16 \times 10^{-5}$ \\
        $\mathcal{M}_2$ & 3 & 3 & 8 & 0.1 & 2000 & 75300& $1.13 \times 10^{-5}$ & $4.31 \times 10^{-6}$ & $4.62 \times 10^{-5}$ \\
        \bottomrule
    \end{tabular}
\end{table*}

\begin{figure*}[htp!]
    \centering
    \begin{subfigure}[t]{\picWidth}
        \centering
        \begin{tikzpicture}
  \begin{axis}[
    width=0.9\linewidth,
    height=6cm,
    xlabel={Number of moments},
    ylabel={$L^2$ Loss},
    grid=both,
    ymode=log,
    ymin=1e-7,
    ymax=0, 
    xtick={1,2,3,4,5,6,7},
    ytickten={-7,-6,-5,-4,-3,-2, -1, 0}, 
    legend style={at={(0.5,-0.2)}, anchor=north, legend columns=2},
    legend cell align={left}
  ]

    \addplot[
      only marks,
      mark=*,
      blue,
      mark size=2pt
    ] coordinates {
      (1, 0.3129636276)
      (2, 0.0363436191)
      (3, 0.0363277850)
      (4, 0.0009605165)
      (5, 0.0009096088)
      (6, 0.0001387931)
    };
    \addlegendentry{Legendre decomp.}

        \addplot[
      only marks,
      mark=*,
      orange,
      mark size=2pt
    ] coordinates {
      (1, 0.04151615908599297)
      (2, 0.002278241201107986)
      (3, 0.0017718592103195942)
      (4, 0.0005890159904020099)
      (5, 0.0036366118508840656)
      (6, 0.002258842833525505)
    };
    \addlegendentry{AE-POD}

     \addplot[
      only marks,
      mark=*,
      green,
      mark size=2pt
    ] coordinates {
      (1, 0.12)
      (2, 0.0030)
      (3, 0.00059)
      (4, 0.000062)
      (5, 0.000021)
      (6, 0.0000063)
      
    };
    \addlegendentry{POD}

    \addplot[
      red,
      thick,
      dashed
    ] coordinates {
      (1, 0.009033438)
      (6, 0.009033438)
    };
    \addlegendentry{IDEA loss = 0.0090}

  \end{axis}
\end{tikzpicture}  
        \caption{$\mathcal{M}_1$ results for each models up to 6 parameters}
        \label{fig:11}
    \end{subfigure}
    ~
    \begin{subfigure}[t]{\picWidth}
        \centering
        \begin{tikzpicture}
  \begin{axis}[
    width=0.9\linewidth,
    height=6cm,
    xlabel={Number of moments},
    ylabel={$L^2$ Loss},
    grid=both,
    ymin=1e-5,
    ymode=log,
    xtick={1,2,3,4,5,6,7},
    legend style={at={(0.5,-0.2)}, anchor=north, legend columns=2},
    legend cell align={left}
  ]

    \addplot[
      only marks,
      mark=*,
      blue,
      mark size=2pt
    ] coordinates {
      (1, 0.4800108592848015)
      (2, 0.04531730240561404)
      (3, 0.0333595880678947)
      (4, 0.021674706069265574)
      (5, 0.014910301515957946)
      (6, 0.009570089901546088)
      (7, 0.0065268596013871695)

    };
    \addlegendentry{Legendre decomp.}

        \addplot[
      only marks,
      mark=*,
      green,
      mark size=2pt
    ] coordinates {
      (1, 0.081)
      (2, 0.036)
      (3, 0.013)
      (4, 0.0070)
      (5, 0.0029)
      (6, 0.0016)
      (7, 0.00085)

    };
    \addlegendentry{POD}

    \addplot[
      only marks,
      mark=*,
      orange,
      mark size=2pt
    ] coordinates {
      (1, 0.038)
      (2, 0.00916)
      (3, 0.0021)
      (4, 0.0020)
      (5, 0.0010)
      (6, 0.0030)
      (7, 0.0018)

    };
    \addlegendentry{AE-POD}

    \addplot[
      red,
      thick,
      dashed
    ] coordinates {
      (1, 0.0065268596013)
      (7, 0.0065268596013)
    };
    \addlegendentry{IDEA loss = 0.0065}

  \end{axis}
\end{tikzpicture}
        \caption{$\mathcal{M}_2$ results for each models up to 7 parameters}
        \label{fig:12}
    \end{subfigure}
    \caption{Comparison of the test-set relative $L_2$ reconstruction loss for IDEA, POD, and AE-POD, against projection onto the first $n$ scaled Legendre polynomials.}
    \label{fig:22}
\end{figure*}

As observed in \cref{tab:tl3}, IDEA predicts $\tilde d = 3$ intrinsic dimensions, including one latent variable for the water height. This correctly suggests that the velocity profiles lie on a 2-dimensional manifold, as expected.
Fig. \ref{fig:corr2} presents the correlation between the 2-dimensional mapping of the full solution and the projection coefficient of the velocity profiles on the scaled Legendre basis computed $\forall x, t$.

As expected, the first latent variable is perfectly correlated with the water height $h$. Furthermore, $L_1$ is perfectly negatively correlated with $\alpha_0$ and $L_2$ is a combination of $\alpha_1$ and $\alpha_3$. The correlation between $\alpha_1$ and $\alpha_3$ is captured by our model. IDEA needs only one variable to simulate the action of 
$\alpha_1$ and $\alpha_3$, whereas a moment model would need both projection coefficients. This means that IDEA indicates the possibility of a more efficient representation using a non-linear ansatz. 

Fig. \ref{fig:22} depicts the relative $L2$ loss of the projection of the reference solution on a subset $\mathcal{S}_n = \{ P_1, \ldots, P_n \}$ of the scaled Legendre polynomial basis.
This truncation loss is compared with the loss of IDEA on the test set. We see that for $\mathcal{M}_1$, 4 moments are necessary to make the projection more accurate than the 2-dimensional mapping of IDEA. This shows that a trivial moment model with dense basis is not efficient in approximating the data, while the sparse representation with only 2 latent variables for the velocity field as computed by IDEA is both accurate and efficient. 

Note that the relative $L2$ loss is computed as
\begin{equation}
    \text{Loss} = \frac{\lVert X - \tilde X \rVert_F}{\lVert X \rVert_F},
\end{equation}
where $X$ is the true dataset, $\tilde X$ is the prediction and $\lVert \cdot \rVert_F$ the Frobenius norm.

\subsubsection{Second case: complex solution with a large number of moments ($\mathcal{M}_2$)}

\begin{figure}[htp!]
\setlength\figureheight{0.2\textwidth}
	\setlength\figurewidth{0.2\textwidth}
	\centering
    \scalebox{0.75}{\begin{tikzpicture}
  \begin{axis}[
    width=\linewidth,
    height=6cm,
    xlabel={t},
    ylabel={$\alpha_i$},
    legend style={at={(1.05,1)}, anchor=north west},
    grid=both,
    thick,
    cycle list name=color list
  ]

    \addplot table [x=time, y={Alpha 2}, col sep=comma] {body/figures/moments_2.csv};
    \addlegendentry{$\alpha_2$}
    
    \addplot table [x=time, y={Alpha 3}, col sep=comma] {body/figures/moments_2.csv};
    \addlegendentry{$\alpha_3$}
    
    \addplot table [x=time, y={Alpha 4}, col sep=comma] {body/figures/moments_2.csv};
    \addlegendentry{$\alpha_4$}

    \addplot table [x=time, y={Alpha 5}, col sep=comma] {body/figures/moments_2.csv};
    \addlegendentry{$\alpha_5$}

    \addplot table [x=time, y={Alpha 6}, col sep=comma] {body/figures/moments_2.csv};
    \addlegendentry{$\alpha_6$}

    \addplot table [x=time, y={Alpha 7}, col sep=comma] {body/figures/moments_2.csv};
    \addlegendentry{$\alpha_7$}

    \addplot table [x=time, y={Alpha 8}, col sep=comma] {body/figures/moments_2.csv};
    \addlegendentry{$\alpha_8$}

    \addplot table [x=time, y={Alpha 9}, col sep=comma] {body/figures/moments_2.csv};
    \addlegendentry{$\alpha_9$}

  \end{axis}
\end{tikzpicture}}
    \caption{Time evolution of coefficients $\alpha_i, i=2,\ldots,9$, averaged along x. As $\alpha_0$ and $\alpha_1$ are on larger scales, they are excluded from the plot to avoid distortion.}
    \label{fig:frt2}
\end{figure}
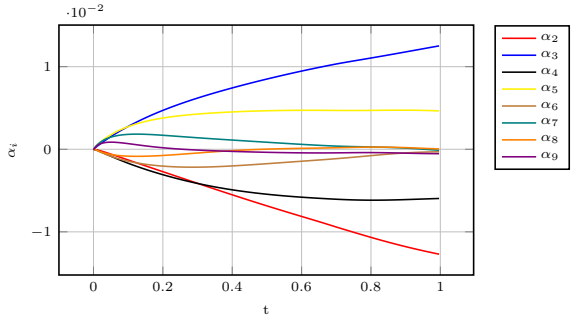

For the second test case, we set $\chi = 0.01$ and $R = 0.01$, these coefficients create velocity profiles with a complex dependency on $\zeta$, involving several moments, see \cite{TorrilhonKowalski}.
Unlike case 1, the moments do not stabilize and $|\alpha_i|$ is increasing with time for the majority of moments, meaning that the velocity profile becomes increasingly complex, as shown in Fig. \ref{fig:frt2}.

As \cref{tab:tl3} indicates, IDEA predicts an intrinsic dimension $\tilde{d}=3$, including keeping one latent variable for the water height. As expected, the model correctly estimates that these complex velocity profiles lie in a 2-dimensional manifold.
This result indicates that while a moment model would need more parameters to describe more complex solutions, IDEA captures underlying dependencies that allow the reconstruction of the profiles using only two dimensions, the correctly estimated intrinsic dimension of the solution.

\begin{figure}[htp!]
\centering
\scalebox{0.9}{
\begin{tikzpicture}[scale=0.6]

  \shade[top color=myviolet!99!lightgray,bottom color=lightgray]
      (14,-13.5) rectangle (14.5,-0.5);
  \draw[gray] (14,-13.5) rectangle (14.5,-0.5);

  \node[anchor=west] at (14.5,-0.5) {1};
  \node[anchor=west] at (14.5,-13.5) {0};

  \foreach \y [count=\n] in {
{ 1.0, -0.0, -0.0,  1.0,  0.0, -0.6,  0.2,  0.3,  0.2,  0.2,  0.1, 0.0, -0.1},
{-0.0,  1.0, -0.0, -0.0, -0.8,  0.2,  0.6,  0.6,  0.7,  0.7,  0.6, 0.5,  0.3},
{-0.0, -0.0,  1.0, -0.0, -0.0,  0.5, -0.0, -0.2, -0.2, -0.2, -0.2, -0.1, -0.1},
{ 1.0, -0.0, -0.0,  1.0,  0.0, -0.6,  0.2,  0.4,  0.2,  0.2,  0.1, 0.0, -0.1},
{ 0.0, -0.8, -0.0,  0.0,  1.0, -0.3, -0.7, -0.7, -0.9, -0.9, -0.7, -0.6, -0.4},
{-0.6,  0.2,  0.5, -0.6, -0.3,  1.0,  0.0, -0.1, -0.0, -0.0, -0.0, -0.0, -0.0},
{ 0.2,  0.6, -0.0,  0.2, -0.7,  0.0,  1.0,  0.9,  0.8,  0.6,  0.3, 0.1,  0.0},
{ 0.3,  0.6, -0.2,  0.4, -0.7, -0.1,  0.9,  1.0,  0.8,  0.8,  0.4, 0.2,  0.0},
{ 0.2,  0.7, -0.2,  0.2, -0.9, -0.0,  0.8,  0.8,  1.0,  0.9,  0.7, 0.4,  0.2},
{ 0.2,  0.7, -0.2,  0.2, -0.9, -0.0,  0.6,  0.8,  0.9,  1.0,  0.8, 0.6,  0.3},
{ 0.1,  0.6, -0.2,  0.1, -0.7, -0.0,  0.3,  0.4,  0.7,  0.8,  1.0, 0.7,  0.5},
{ 0.0,  0.5, -0.1,  0.0, -0.6, -0.0,  0.1,  0.2,  0.4,  0.6,  0.7, 1.0,  0.6},
{-0.1,  0.3, -0.1, -0.1, -0.4, -0.0,  0.0,  0.0,  0.2,  0.3,  0.5, 0.6,  1.0},
    } {

      \foreach \x [count=\m] in \y {

        \pgfmathsetmacro{\val}{abs(\x)*100}
        \node[
          fill=myviolet!\val!lightgray, 
          minimum width=6mm, minimum height=6mm,
          text width=6mm,
          align=center,
          text=White,
          font=\footnotesize,
          inner sep=0pt,
          outer sep=0pt,
          draw=gray
        ] at (\m,-\n) {\x};
      }
    }

  \foreach \a [count=\i] in {$L_0$,$L_1$,$L_2$,$h$,$\alpha_0$,$\alpha_1$,$\alpha_2$, $\alpha_3$, $\alpha_4$, $\alpha_5$, $\alpha_6$, $\alpha_7$, $\alpha_8$} {
    \node[minimum size=6mm] at (0,-\i) {\a};
  }

  \foreach \label [count=\i from 1] in {$L_0$,$L_1$,$L_2$,$h$,$\alpha_0$,$\alpha_1$,$\alpha_2$, $\alpha_3$, $\alpha_4$, $\alpha_5$, $\alpha_6$, $\alpha_7$, $\alpha_8$} {
  \node[minimum size=6mm] at (\i, 0) {\label};
}
\end{tikzpicture}}
\caption{Correlation matrix between the three latent variables $L_i$ and the coefficients $a_i$. The color varies according to the absolute value of the correlation between the parameters.}
\label{fig:corr3}
\end{figure}
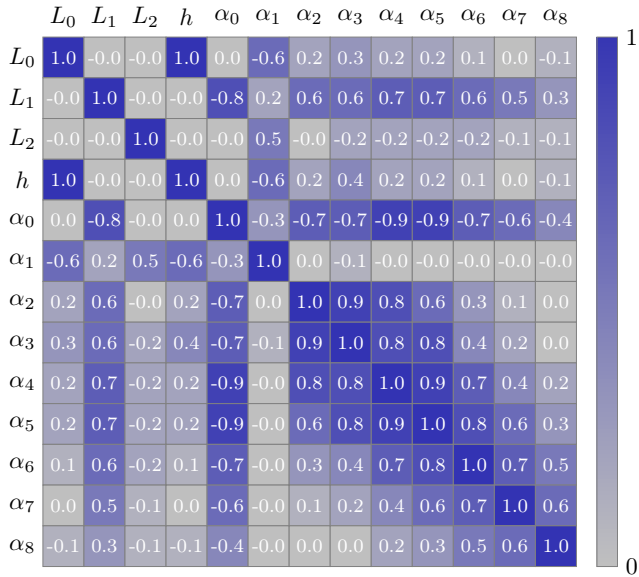

Correlations are shown in Fig. \ref{fig:corr3}. Consistent with expectations, the first latent variable is perfectly correlated with the water height $h$. 
The two other latent variables now learn a combination of all remaining moments of the solution, leveraging the fact that these moments are correlated. This level of compression is difficult to achieve with traditional moment models, which require more parameters to reach comparable accuracy. In Fig. \ref{fig:22} (b), for example, it is shown that at least 7 coefficients $\alpha_i$ are necessary to capture the data to a comparable accuracy, while IDEA is able to achieve this accuracy only using 2 latent variables.

In sum, IDEA successfully predicts the intrinsic dimension of the two application test datasets and succeeds at accurate reconstructions. 

\subsection{Comparison with other Model Order Reduction methods.}

We compare IDEA's results with two other Model Order Reduction methods.

\subsubsection{Proper Orthogonal Decomposition (POD)}

We compared IDEA with Proper Orthogonal Decomposition (POD) of the vertical velocity profiles. Specifically, we applied PCA to the $n_{\zeta}$ vertical levels, treating each profile as a vector in $\mathbb{R}^{n_{\zeta}}$. PCA identifies a low-dimensional linear subspace that captures most of the data variance, and represents each profile by its coordinates in the resulting orthogonal basis. By construction, POD provides the optimal linear $L^2$ reconstruction among all $k$-dimensional bases. Fig. \ref{fig:corr4} shows that the first vectors of that base are close to the Legendre polynomial basis. Fig. ~\ref{fig:22} reports POD results on both flow datasets: it matches IDEA on the simple case with only 2 modes, but requires 5 modes to reach IDEA's accuracy on the more complex (nonlinear) case. Finally, POD modes may be harder to interpret physically since they are purely data-driven.

\subsubsection{Non-linear POD extension.}

We additionally benchmark IDEA against a non-linear extension of POD: an autoencoder-based ROM (AE-POD), following the nonlinear-manifold ROM literature (e.g., \cite{Ahmed_2021}). Reconstruction losses are reported in Table~\ref{tab:tl3}. Fig. ~\ref{fig:corr4} further illustrates the correlation between the learned latent variables and the physically meaningful Legendre coefficients for the $\mathcal{M}_2$ case.

\begin{figure}[htp!]
\centering
\scalebox{0.9}{
\begin{tikzpicture}[scale=0.6]

  \shade[top color=myviolet!99!lightgray,bottom color=lightgray]
      (9.6,-7.5) rectangle (10.1,-0.5);
  \draw[gray] (9.6,-7.5) rectangle (10.1,-0.5);

  \node[anchor=west] at (10.1,-0.5) {1};
  \node[anchor=west] at (10.1,-7.5) {0};

  \foreach \y [count=\n] in {
    { 0.7, -1.0,  0.3,  0.4,  1.0, -0.1,  0.6, -0.6},
    {-0.7,  0.4,  0.8,  0.7,  0.0, -1.0,  0.0,  0.1},
    {-0.1,  0.8, -0.3, -0.6,  0.2,  0.4,  0.8,  0.2},
    {-0.2,  0.7, -0.5, -0.8,  0.2,  0.7,  0.6,  0.0},
    {-0.3,  0.9, -0.5, -0.7, -0.2,  0.7,  0.0, -0.3},
    {-0.5,  0.8, -0.6, -0.6, -0.5,  0.6, -0.7, -0.1},
    {-0.5,  0.7, -0.5, -0.4, -0.7,  0.3, -0.9,  0.2},
  }{
    \foreach \x [count=\m] in \y {
      \pgfmathsetmacro{\val}{min(100,abs(\x)*100)}
      \node[
        fill=myviolet!\val!lightgray,
        minimum width=6mm, minimum height=6mm,
        text width=6mm,
        align=center,
        text=white,
        font=\footnotesize,
        inner sep=0pt,
        outer sep=0pt,
        draw=gray
      ] at (\m,-\n) {\x};
    }
  }

  \foreach \a [count=\i] in {$\alpha_0$,$\alpha_1$,$\alpha_2$,$\alpha_3$,$\alpha_4$,$\alpha_5$,$\alpha_6$} {
    \node[anchor=east, minimum size=6mm] at (0.2,-\i) {\a};
  }


\node[anchor=south, font=\scriptsize] at (2.6,0.2) {AE-POD};
\node[anchor=south, font=\scriptsize] at (6.5,0.2) {POD};

\foreach \j in {1,2,3,4} {
  \node[anchor=south, font=\scriptsize] at (\j,-0.40) {\j};
  \node[anchor=south, font=\scriptsize] at (\j+4,-0.40) {\j};
}
\end{tikzpicture}}
\caption{Absolute Pearson correlation between the four latent variables learned by the AE-POD$_4$ model and the POD$_4$ coefficients, for the Legendre coefficients $\{\alpha_0,\dots,\alpha_6\}$. The color varies according to the absolute value of the correlation between the parameters.}
\label{fig:corr4}
\end{figure}

As shown in Fig.~\ref{fig:corr4}, although AE-POD is slightly more accurate in terms of reconstruction, its latent variables are less interpretable than POD. In contrast, IDEA achieves comparable accuracy while better preserving physical structure, as evidenced by the clearer correlations with the Legendre coefficients in Fig.~\ref{fig:corr3}. Regarding computational efficiency, POD is indeed significantly faster (closed-form SVD), whereas AE-POD has a training cost comparable to IDEA due to the similar neural-network optimization procedure.

\section{Conclusion}
We introduced IDEA, the Intrinsic Dimension Estimating Autoencoder, as
a neural network autoencoder model designed to estimate the
intrinsic dimension of datasets while preserving strong reconstruction capabilities.
By leveraging the flexible architecture offered by the autoencoder architecture and a novel regularization scheme through an innovative use of re-weighted CancelOut layers, IDEA identifies and suppresses redundant latent variables directly within the network, thanks to a projected loss and latent space disentanglement, effectively identifying dimensional 
structure underlying the data.

Through a series of synthetic and theoretical benchmarks, IDEA demonstrated competitive performance compared
to state-of-the-art intrinsic dimension estimators. Unlike classical
estimators, our model directly provides integer-valued intrinsic dimensions by counting active latent variables. This 
offers an interpretable and adjustable latent space on which analysis can be directly performed, thus enabling more efficient computation. Additionally, IDEA provides a direct reconstruction of the data.

Applying IDEA to a physically motivated case such as the one-dimensional free-surface flow, our model successfully identified a low-dimensional latent representation that preserved the key dynamical structures of the system.
We showed that IDEA is a significant advantage over standard moment models, which required a larger number of parameters to capture similar dynamics, due to their limited ability to exploit correlations between higher-order moments.

\addcontentsline{toc}{section}{Data and Code availability}
\section*{Data and Code availability}
Code is available on Zenodo:  \href{https://doi.org/10.5281/zenodo.18314616}{https://doi.org/10.5281/zenodo.18314616}.

\addcontentsline{toc}{section}{Author Contribution Statement (CRediT)}
\section*{Author Contribution Statement (CRediT)}
\textbf{Antoine Oriou:} Conceptualisation, Methodology, Software, Investigation, Numerical Experiments, Writing - Original Draft, Visualisation\\
\textbf{Philipp Krah:} Conceptualisation, Methodology, Writing - Original Draft, Supervision, Writing - Review \& Editing\\
\textbf{Julian Koellermeier:} Conceptualisation, Methodology, Writing - Original Draft, Supervision, Writing - Review \& Editing

\addcontentsline{toc}{section}{Acknowledgement}
\section*{Acknowledgement}
This publication is part of the project \textit{HiWAVE} with file number VI.Vidi.233.066 of the \textit{ENW Vidi} research programme, funded by the \textit{Dutch Research Council (NWO)}.

\appendix

\section{Model reconstruction and accuracy on benchmark datasets}
\label{sec:BM}

As our model is able to reconstruct the dataset under the estimated dimension, we can compute reconstruction losses. 
As stated before, when the predicted dimension is $\tilde d$, the model keeps track of the best reconstruction estimator with $\tilde d+1$, $\tilde d$ and $\tilde d-1$ latent dimensions (see benchmark results in \cref{tab:tl2}).
We trained the model on all the datasets for a total of 1000 epochs, reserving 20\% of the data for testing. 
The weights of the CancelOut layers were initialized to $1$, 
while all linear layers use PyTorch's default initialization 
(Kaiming uniform). 
The \textit{best model} at each candidate latent dimensionality 
is selected dynamically during training by monitoring the test reconstruction and projected loss. During training a dictionary of the best states is maintained:
It saves the model state for the current $l_\text{eff}$ dimension if its test loss improves. It stores an $l_\text{eff} - 1$ version with the last CancelOut weight forced to zero if this gives a lower projected loss. It prunes states corresponding to dimensions that are $\ge l_\text{eff} + 2$. This ensures that for every plausible latent dimension $\tilde d$, we retain the optimal set of parameters found during training.

\begin{figure*}[htbp!]
  
    \centering
 
    \includegraphics[width=1\linewidth]{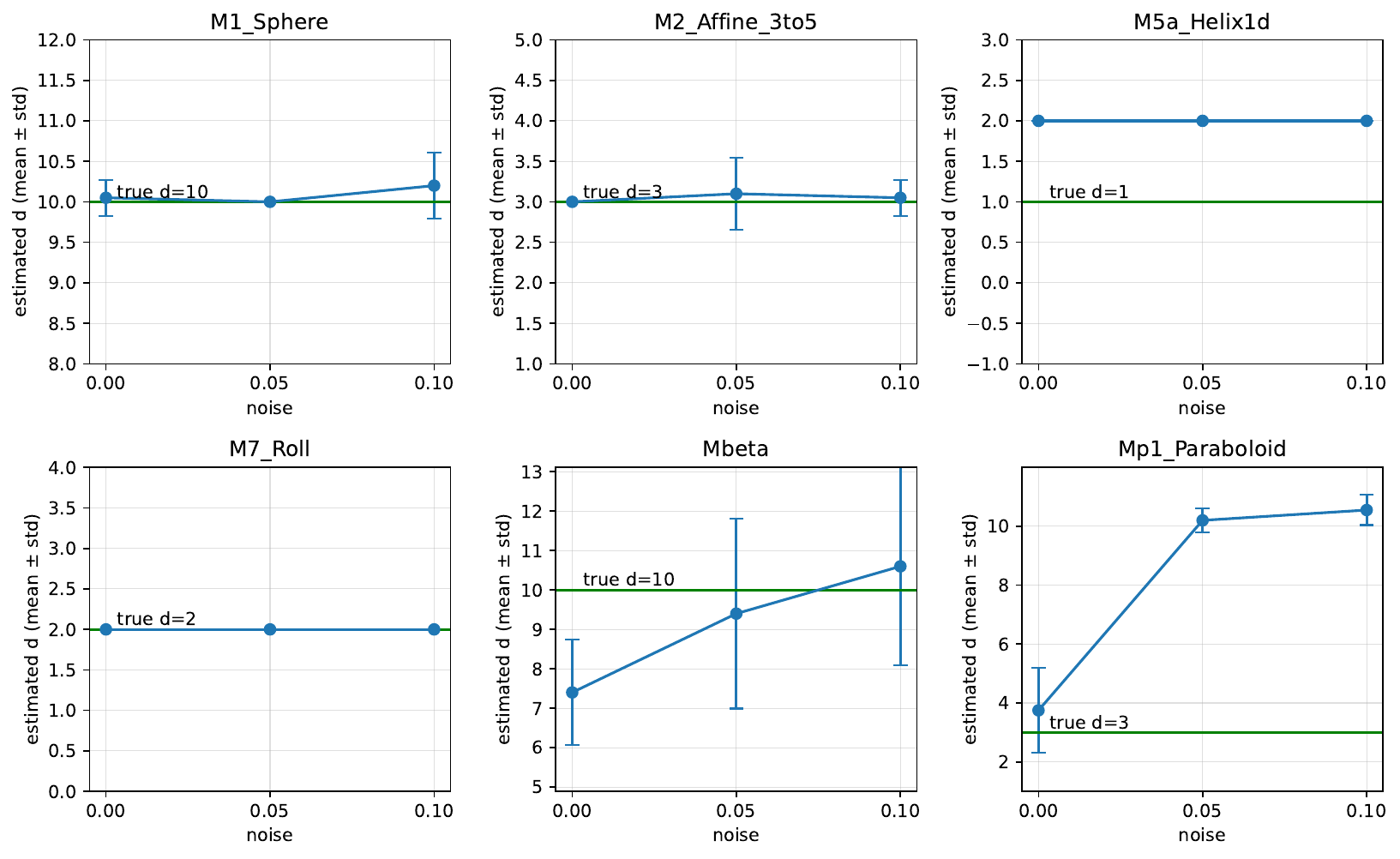}
    \caption{Intrinsic-dimension estimation robustness to noise. For each dataset and noise level, IDEA was trained 20 times (different random seeds) on normalized samples with additive noise. Markers show the mean predicted intrinsic dimension $\hat d$ across runs, vertical error bars indicate $\pm$ 1 standard deviation, and the horizontal line denotes the ground-truth intrinsic dimension d.}

    \label{fig:noise}
\end{figure*}

Additionally, we ran IDEA multiple times to assess the reliability of its intrinsic-dimension estimates under increasing levels of additive noise. For each noise level, we performed 20 independent trainings and report the mean predicted dimension with error bars (standard deviation), together with the ground-truth intrinsic dimension. Noise is added by perturbing each sample with i.i.d. Gaussian noise $\varepsilon \sim \mathcal{N}(0,\texttt{noise})$, where $\texttt{noise}$ denotes the noise parameter. Results for dataset TF are shown in Fig.~\ref{fig:noise}. We see that IDEA is robust to noise for the major part of the dataset, except for $\texttt{Mp1\_Paraboloid}$.

\begin{table*}[htbp!]
    \centering
    \caption{Summary of reconstruction error across datasets and projected dimensions. We present in this table the most accurate model with the smaller size of dataset. $l$ is the initial number of non-zero weights of the model. The columns $\tilde d$, \textbf{\# Time}, \textbf{$\mathcal{L}_{\text{test}}$} are averaged on 5 independent trainings (except for $\mathcal{M}_6$ which is averaged on 2 independent trainings). Standard deviations are reported with the $\pm$ notation. Reported computation times correspond to single-thread execution without GPU acceleration.}
    \label{tab:tl2}
    \small
    \setlength{\tabcolsep}{6pt}
    \renewcommand{\arraystretch}{1.2}
    \begin{tabular}{l c c c c c c c c}
        \toprule
        \textbf{Dataset} 
        & \textbf{$d$} 
        & \textbf{$\tilde{d}$} 
        & \textbf{$l$}
        & \textbf{\# Samples}
        & \textbf{\# Time}
        & \multicolumn{3}{c}{\textbf{$\mathcal{L}_{\text{test}}$}} \\
        \cmidrule(lr){7-9}
        & & & & & &  $\tilde d+1$ & $\tilde d$ & $\tilde d-1$ \\
        \midrule
        $\mathcal{M}_1$ & 10 & 10.0 & 11 & 20000 & 3min51s  &
        $2.9\mathrm{e}{-4}\,\pm\,4.9\mathrm{e}{-6}$  &
        $3.0\mathrm{e}{-4}\,\pm\,1.7\mathrm{e}{-4}$  &
        $4.4\mathrm{e}{-3}\,\pm\,2.6\mathrm{e}{-3}$ \\

        $\mathcal{M}_2$ & 3 & 3.0 & 5 & 15000 & 1min51s  &
        $5.3\mathrm{e}{-6}\,\pm\,7.3\mathrm{e}{-7}$ &
        $5.3\mathrm{e}{-6}\,\pm\,7.1\mathrm{e}{-7}$ &
        $4.0\mathrm{e}{-4}\,\pm\,4.7\mathrm{e}{-5}$ \\

        $\mathcal{M}_3$ & 4 & 5.0 & 6 & 20000 & 3min47s  &
        $1.9\mathrm{e}{-5}\,\pm\,1.4\mathrm{e}{-5}$ &
        $1.9\mathrm{e}{-5}\,\pm\,5.1\mathrm{e}{-7}$ &
        $1.0\mathrm{e}{-4}\,\pm\,2.0\mathrm{e}{-6}$ \\

        $\mathcal{M}_4$ & 4 & 4.0 & 8 & 75000 & 13min37s &
        $1.9\mathrm{e}{-4}\,\pm\,1.6\mathrm{e}{-4}$ &
        $2.0\mathrm{e}{-4}\,\pm\,2.3\mathrm{e}{-4}$ &
        $5.5\mathrm{e}{-3}\,\pm\,1.9\mathrm{e}{-4}$ \\

        $\mathcal{M}_{5a}$ & 1 & 2.0 & 3 & 20000 & 3min41s  &
        $3.7\mathrm{e}{-6}\,\pm\,3.5\mathrm{e}{-6}$ &
        $3.9\mathrm{e}{-6}\,\pm\,5.0\mathrm{e}{-6}$ &
        $4.2\mathrm{e}{-3}\,\pm\,1.2\mathrm{e}{-3}$ \\

        $\mathcal{M}_{5b}$ & 2 & 2.0 & 3 & 15000 & 1min42s  &
        $5.3\mathrm{e}{-6}\,\pm\,1.9\mathrm{e}{-6}$ &
        $5.9\mathrm{e}{-6}\,\pm\,1.8\mathrm{e}{-6}$ &
        $1.7\mathrm{e}{-2}\,\pm\,1.1\mathrm{e}{-3}$ \\

        $\mathcal{M}_6$ & 6 & 6.0 & 16 & 200000 & 37min06s &
        $1.1\mathrm{e}{-4}\,\pm\,2.3\mathrm{e}{-5}$ &
        $1.6\mathrm{e}{-4}\,\pm\,5.6\mathrm{e}{-5}$ &
        $7.0\mathrm{e}{-4}\,\pm\,2.3\mathrm{e}{-4}$ \\

        $\mathcal{M}_7$ & 2 & 2.0 & 3 & 15000 & 2min15s  &
        $4.4\mathrm{e}{-6}\,\pm\,8.1\mathrm{e}{-7}$ &
        $4.4\mathrm{e}{-6}\,\pm\,1.2\mathrm{e}{-6}$ &
        $2.0\mathrm{e}{-2}\,\pm\,2.0\mathrm{e}{-3}$ \\

        $\mathcal{M}_{beta}$ & 10 & 10.0 & 16 & 15000 & 3min03s  &
        $9.5\mathrm{e}{-5}\,\pm\,1.4\mathrm{e}{-5}$ &
        $9.5\mathrm{e}{-5}\,\pm\,2.3\mathrm{e}{-5}$ &
        $4.0\mathrm{e}{-4}\,\pm\,6.1\mathrm{e}{-5}$ \\

        $\mathcal{M}_{P3}$ & 3 & 3.0 & 12 & 20000 & 5min34s  &
        $4.6\mathrm{e}{-6}\,\pm\,6.6\mathrm{e}{-7}$ &
        $4.6\mathrm{e}{-6}\,\pm\,8.8\mathrm{e}{-7}$ &
        $8.0\mathrm{e}{-5}\,\pm\,3.9\mathrm{e}{-4}$ \\

        $\mathcal{M}_{P6}$ & 6 & 6.0 & 16 & 50000 & 10min36s  &
        $1.6\mathrm{e}{-5}\,\pm\,1.5\mathrm{e}{-6}$ &
        $1.6\mathrm{e}{-5}\,\pm\,1.4\mathrm{e}{-6}$ &
        $3.0\mathrm{e}{-5}\,\pm\,1.4\mathrm{e}{-4}$ \\
        \bottomrule
    \end{tabular}
\end{table*}

Fig. \ref{fig:training_results} shows the evolution of the predicted dimension with the sample size of the dataset.

\begin{figure*}[htbp!]
\centering
\setlength\figurewidth{0.3\textwidth}
\scalebox{1}{
\begin{tikzpicture}
\begin{axis}[
    width=16cm,
    height=9cm,
    xtick={1,...,11},
    xticklabels={
      M1\_Sphere,
      M2\_Affine\_3to5,
      M3\_Nonlin\_4to6,
      M4\_Nonlin,
      M5a\_Helix1d,
      M5b\_Helix2d,
      M6\_Nonlin,
      M7\_Roll,
      Mbeta,
      Mp1\_Paraboloid,
      Mp2\_Paraboloid},
    xticklabel style={rotate=45,anchor=east},
    ylabel={Predicted intrinsic dimension},
    legend pos=outer north east,
    ymajorgrids=true,
    grid style=dashed,
]

\addplot+[only marks,
    mark=-,
    mark options={red, ultra thick},
    mark size=8pt]
coordinates {
  (1,10) (2,3) (3,4) (4,4) (5,1) (6,2) (7,6) (8,2) (9,10) (10,3) (11,6)
};
\addlegendentry{True ID}

\pgfplotsset{myerr/.style={
    error bars/.cd,
      y dir=both,
      y explicit,
}}

\addplot+[only marks, mark=o, color=red, myerr]
coordinates {
  (0.75,10.17) +- (0,0.39)
  (1.75, 3.00) +- (0,0.00)
  (2.75, 5.00) +- (0,0.00)
  (3.75, 7.00) +- (0,0.00)
  (4.75, 2.00) +- (0,0.00)
  (5.75, 2.00) +- (0,0.00)
  (6.75,11.73) +- (0,1.01)
  (7.75, 2.00) +- (0,0.00)
  (8.75,10.00) +- (0,0.50)
  (9.75, 3.17) +- (0,0.39)
  (10.75,6.22) +- (0,0.43)
};
\addlegendentry{15k samples}

\addplot+[only marks, mark=square, color=blue, myerr]
coordinates {
  (0.85,10.33) +- (0,0.58)
  (1.85, 3.00) +- (0,0.00)
  (2.85, 5.00) +- (0,0.00)
  (3.85, 7.00) +- (0,0.00)
  (4.85, 2.00) +- (0,0.00)
  (5.85, 2.00) +- (0,0.00)
  (6.85,11.20) +- (0,0.80)
  (7.85, 2.00) +- (0,0.00)
  (9.85, 3.00) +- (0,0.20)
  (10.85,6.20) +- (0,0.30)
};
\addlegendentry{20k samples}

\addplot+[only marks, mark=triangle, color=green, myerr]
coordinates {
  (2.95,5.00) +- (0,0.00)
  (3.95,4.00) +- (0,0.00)
  (4.95,2.00) +- (0,0.00)
  (6.95,10.70) +- (0,0.70)
  (10.95,6.00) +- (0,0.20)
};
\addlegendentry{50k samples}

\addplot+[only marks, mark=diamond, color=orange, myerr]
coordinates {
  (3.05,5.00) +- (0,0.00)
  (4.05,4.00) +- (0,0.00)
  (5.05,2.00) +- (0,0.00)
  (7.05,10.30) +- (0,0.70)
};
\addlegendentry{75k samples}

\addplot+[only marks, mark=star, color=purple, myerr]
coordinates {
  (3.15,5.00) +- (0,0.00)
  (5.15,2.00) +- (0,0.00)
  (7.15,8.30) +- (0,0.60)
};
\addlegendentry{150k samples}

\addplot+[only marks, mark=*, color=brown, myerr]
coordinates {
  (3.25,5.00) +- (0,0.00)
  (5.25,2.00) +- (0,0.00)
  (7.25,6.00) +- (0,0.40)
};
\addlegendentry{200k samples}

\end{axis}
\end{tikzpicture}}
\caption{True intrinsic dimension $d$ and evolution of IDEA’s predicted dimension  $\tilde d$ across benchmark datasets, trained with varying sample sizes. Markers show the mean over repeated trainings, and error bars denote $\pm$ one standard deviation. Results are averaged on 5 independent trainings (except for sizes 150k and 200k, which are averaged on 2 independent trainings).}
    \label{fig:training_results}
\end{figure*}

\bibliographystyle{elsarticle-harv}
\bibliography{bibliography.bib}

\end{document}